\documentclass[12pt]{article}
\usepackage[letterpaper,margin=2cm]{geometry}
\usepackage{natbib}
\usepackage{algorithm2e}
\usepackage{algorithmic}
\usepackage{amsmath}
\usepackage{amssymb}
\usepackage{booktabs}
\usepackage{caption}
\usepackage{subcaption}
\usepackage{graphicx}
\usepackage{amsfonts}
\usepackage{hyperref}

\newcommand{\Tau}{\mathcal{T}}
\newcommand{\loss}{\mathcal{L}}

\newcommand{\hid}{\mathbf{h}}
\newcommand{\inp}{\mathbf{x}}
\newcommand{\out}{\mathbf{y}}

\newcommand{\weight}{\mathbf{W}}
\newcommand{\bias}{\mathbf{b}}
\newcommand{\fgate}{\mathbf{f}}
\newcommand{\igate}{\mathbf{i}}
\newcommand{\ogate}{\mathbf{o}}
\newcommand{\cgate}{\bar{\mathbf{c}}}
\newcommand{\cell}{\mathbf{c}}
\newcommand{\act}{\mathbf{a}}

\newcommand{\Dtr}{D^{tr}_{\Tau_j}}
\newcommand{\Dte}{D^{te}_{\Tau_j}}

\DeclareMathOperator*{\expect}{\mathbb{E}}

\title{Are LSTMs Good Few-Shot Learners?\thanks{Accepted at Machine Learning Journal, Special Issue of the ECML PKDD 2023 Journal Track}}

\author{Mike Huisman \and Thomas M. Moerland \and Aske Plaat \and Jan N. van Rijn}

\date{Leiden Institute of Advanced Computer Science, Leiden University}

\begin{document}

\maketitle

\begin{abstract}
Deep learning requires large amounts of data to learn new tasks well, limiting its applicability to domains where such data is available. 
Meta-learning overcomes this limitation by learning how to learn.
In 2001, Hochreiter et al.\ showed that an LSTM trained with backpropagation across different tasks is capable of meta-learning.
Despite promising results of this approach on small problems, and more recently, also on reinforcement learning problems, the approach has received little attention in the supervised few-shot learning setting.
We revisit this approach and test it on modern few-shot learning benchmarks. 
We find that LSTM, surprisingly, outperform the popular meta-learning technique MAML on a simple few-shot sine wave regression benchmark, but that LSTM, expectedly, fall short on more complex few-shot image classification benchmarks. 
We identify two potential causes and propose a new method called \emph{Outer Product LSTM (OP-LSTM)} that resolves these issues and displays substantial performance gains over the plain LSTM.
Compared to popular meta-learning baselines, OP-LSTM yields competitive performance on within-domain few-shot image classification, and performs better in cross-domain settings by 0.5\% to 1.9\% in accuracy score. 
While these results alone do not set a new state-of-the-art, the advances of OP-LSTM are orthogonal to other advances in the field of meta-learning, yield new insights in how LSTM work in image classification, allowing for a whole range of new research directions.
For reproducibility purposes, we publish all our research code publicly.
\end{abstract}

\section{Introduction}

Deep neural networks have demonstrated human or even super-human performance on various tasks in different areas \citep{krizhevsky2012imagenet,he2015delving,mnih2015human,silver2016mastering}.
However, they often fail to learn new tasks well from limited amounts of data \citep{lecun2015deep}, limiting their applicability to domains where abundant data is available. 
\emph{Meta-learning} \citep{brazdil2022metalearning, huisman2021survey, naik1992meta, schmidhuber1987evolutionary, thrun1998lifelong} is one approach to overcome this 
limitation. 
The idea is to learn an efficient learning algorithm over a large number of different tasks so that new tasks can be learned from a few data points. 
Meta-learning involves learning at two different levels: the \emph{inner-level} learning algorithm produces a predictor for the given task at hand, whereas  the \emph{outer-level}  learning algorithm is adjusted to improve  the learning ability across tasks.   

\begin{figure}
    \centering
    \includegraphics[width=\textwidth]{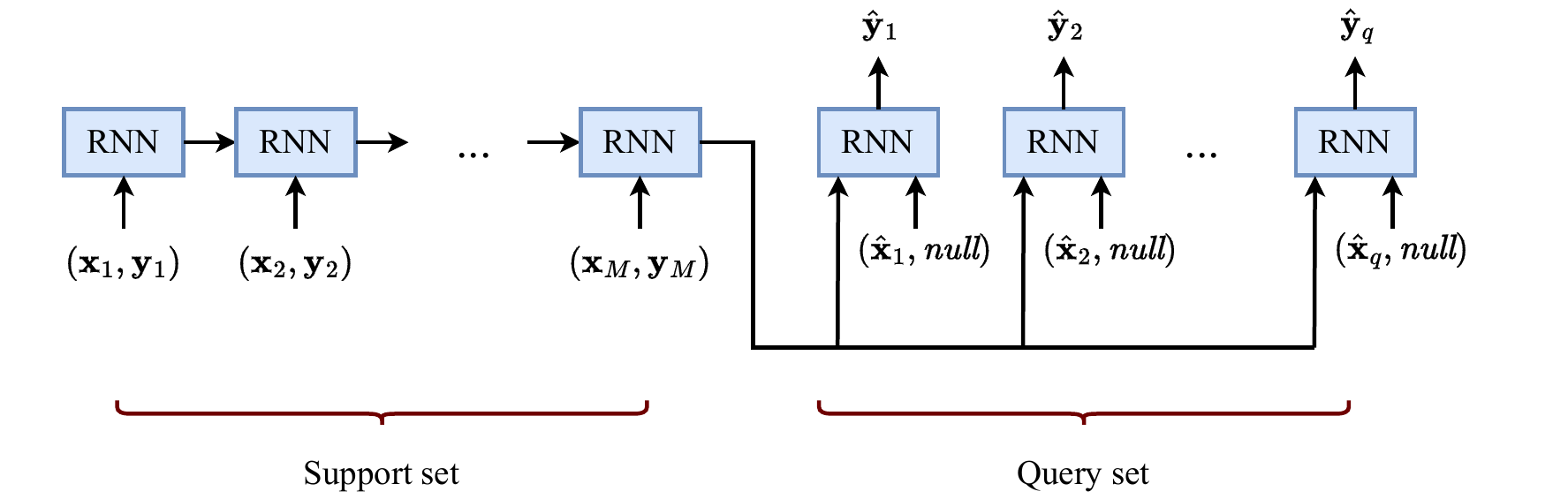}
    \caption{The use of a recurrent neural network for few-shot learning. The support set $\Dtr = \{ (\inp_1, \out_1), \ldots, (\inp_M, \out_M)\}$ is fed as a sequence into the RNN. The predictions $\hat{\out}_j$ for new query points $\hat{x}_j$ are conditioned on the resulting state. We note that feeding the tuples $(\inp_i, \out_i)$ does not lead to the RNN directly outputting the presented labels (drastic overfitting) as the goal is to make predictions for query inputs, for which the ground-truth outputs are unknown. Alternatively, the support set could be fed into the RNN in a temporally offset manner (e.g., feed support tuples $(\inp_{i}, \out_{i-1})$ into the RNN) as in \citet{Santoro16} or in different ways (for example feed the error instead of the ground-truth target) \citep{hochreiter2001learning}.}
    \label{fig:seqprocessing}
\end{figure}

\citet{hochreiter2001learning} and \citet{younger2001meta} have shown that LSTMs trained with gradient descent are capable of meta-learning.
At the inner level---when presented with a new task---the LSTM ingests training examples with corresponding ground-truth outputs and conditions its predictions for new inputs on the resulting hidden state (the general idea for using recurrent neural networks for meta-learning has been visualized in Figure~\ref{fig:seqprocessing}). 
The idea is that the training examples that are fed into the LSTM can be remembered or stored by the LSTM in its internal states, allowing predictions for new unseen inputs to be based on the training examples.
This way, the LSTM can implement a learning algorithm in the recurrent dynamics, whilst the weights of the LSTM are kept frozen. 
During meta-training, the weights of the LSTM are only adjusted at the outer level (across tasks) by backpropagation, which corresponds to updating the inner-level learning program. 
By exposing the LSTM to different tasks which it cannot solve without learning, the LSTM is stimulated to learn tasks by ingesting the training examples which it is fed.
The initial experiments of \citet{hochreiter2001learning} and \citet{younger2001meta} have shown promising results on simple and low-dimensional toy problems. 
Meta-learning with LSTMs has also been successfully extended to reinforcement learning settings \citep{duan2016rl, wang2016learning}, and demonstrates promising learning speed on new tasks.

To the best of our knowledge, the LSTM approach has, in contrast, not been studied on more complex and modern supervised few-shot learning benchmarks by the research community, which has already shifted its attention to more developing new and more complex methods \citep{finn2017model, snell2017prototypical, Flennerhag2020Meta-Learning, parkO19metacurvature}.
In our work, we revisit the idea of meta-learning with LSTMs and study the ability of the learning programs embedded in the weights of the LSTM to perform few-shot learning on modern benchmarks. 
We find that an LSTM outperforms the popular meta-learning technique MAML \citep{finn2017meta} on a simple few-shot sine wave regression benchmark, but that it falls short on more complex few-shot image classification benchmarks.

By studying the LSTM architecture in the context of meta-learning, we identify two potential causes for this underperformance, namely 1)~the fact that it is not invariant to permutations of the training data and 2)~that the input representation and learning procedures are intertwined.
We propose a general solution to the first problem and propose a new meta-learning technique, \emph{Outer Product LSTM (OP-LSTM)}, where we solve the second issue by learning the weight update rule for a base-learner network using an LSTM, in addition to good initialization parameters for the base-learner.
This approach is similar to that of \citet{ravi2017optimization}, but differs in how the weights are updated with the LSTM and that in our approach, the LSTM does not use hand-crafted gradients as inputs in order to produce weight updates.
Our experiments demonstrate that OP-LSTM yields substantial performance gains over the plain LSTM.

Our contributions are the following.
\begin{itemize}
    \item We study the ability of a plain LSTM to perform few-shot learning on modern few-shot learning benchmarks and show that it yields surprisingly good performance on simple regression problems (outperforming MAML \citep{finn2017model}), but is outperformed on more complex classification problems.
    \item We identify two problems with the plain LSTM for meta-learning, namely 1)~the fact that it is not invariant to permutations of the training data and 2)~that the input representation and learning procedures are intertwined, and propose solutions to overcome them by 1)~an average pooling strategy and 2)~decoupling the input representation from the learning procedure.
    \item We propose a novel LSTM architecture called \emph{Outer Product LSTM (OP-LSTM)} that overcomes the limitations of the classical LSTM architecture and yields substantial performance gains on few-shot learning benchmarks. 
    \item We discuss that OP-LSTM can approximate MAML \citep{finn2017model} as well as Prototypical network \citep{snell2017prototypical} as it can learn to perform the same weight matrix updates. Since OP-LSTM does not update the biases, it can only approximate these two methods.
\end{itemize}
Compared to popular meta-learning baselines, including MAML \citep{finn2017model}, Prototypical network \citep{snell2017prototypical}, and Warp-MAML \citep{Flennerhag2020Meta-Learning}, OP-LSTM yields competitive performance on within-domain few-shot image classification, and outperforms them in cross-domain settings by 0.5\% to 1.9\% in raw accuracy score.
While these results alone do not set a new state-of-the-art, the advances of OP-LSTM are orthogonal to other advances in the field of meta-learning, allowing for a whole range of new research directions, such as using OP-LSTM to update the weights in gradient-based meta-learning techniques \citep{Flennerhag2020Meta-Learning, parkO19metacurvature, lee2018gradient} rather than regular gradient descent.
For reproducibility and verifyability purposes, we make all our research code publicly available.\footnote{See: \url{https://github.com/mikehuisman/lstm-fewshotlearning-oplstm}}

\section{Related work}

\paragraph{Earlier work with LSTMs}
Meta-learning with recurrent neural networks was originally proposed by \citet{hochreiter2001learning} and \citet{younger2001meta}.
In their pioneering work, \citet{hochreiter2001learning} also investigated other recurrent neural network architectures for the task of meta-learning, but it was observed that Elman networks and vanilla recurrent neural networks failed to meta-learn simple Boolean functions. 
Only the LSTM was found to be successful at learning simple functions. 
For this reason, we solely focus on LSTM in our work.

The idea of meta-learning with an LSTM at the data level has also been investigated and shown to achieve promising results in the context of reinforcement learning \citep{duan2016rl, wang2016learning, alver2021going}. 
In the supervised meta-learning community, however, the idea of meta-learning with an LSTM at the data level \citep{hochreiter2001learning, younger2001meta} has not gained much attention. 
A possible explanation for this is that \citet{Santoro16} compared their proposed memory-augmented neural network (MANN) to an LSTM and found that the latter was outperformed on few-shot Omniglot \citep{lake2015human} classification.
However, it was not reported how the hyperparameters of the LSTM were tuned and whether it was a single-layer LSTM or a multi-layer LSTM. 
In addition, the LSTM was fed the input data as a sequence which is not permutation invariant, which can hinder its performance. 
We propose a permutation-invariant method of feeding training examples into recurrent neural networks and perform a detailed study of the performance of LSTM on few-shot learning benchmarks.

In concurrent work, \citet{kirsch2022general} investigates the ability of transformer architectures to implement learning algorithms, a baseline with a similar name as our proposed method was proposed (``Outer product LSTM''). 
We emphasize, however, that their method is different from ours (OP-LSTM) as it is a model-based approach that ingests the entire training set and query input into a slightly modified LSTM architecture (with an outer product update and inner product read-out) to make predictions, whereas in our OP-LSTM, the LSTM acts on a meta-level to update the weights of a base-learner network. 

In concurrent works done by \citet{kirsch2022general} and \citet{chan2022data}, the ability of the classical LSTM architecture to implement a learning algorithm was also investigated. 
They observed that it was unable to embed a learning algorithm into its recurrent dynamics on image classification tasks. 
However, the focus was not on few-shot learning, and no potential explanation for this phenomenon was given. 
In our work, we investigate the LSTM's ability to learn a learning algorithm in settings where only one or five examples are present per class, dive into the inner working mechanics to formulate two hypotheses as to why the LSTM architecture is incapable of learning a good learning algorithm, and as a result, propose OP-LSTM which overcomes the limitations and performs significantly better than the classical LSTM architecture.  

\paragraph{Different LSTM architectures for meta-learning} 
\citet{Santoro16} used an LSTM as a read/writing mechanism to an external memory in their MANN technique. 
\citet{kirsch2021meta} proposed to replace every weight in a neural network with a recurrent neural network that communicates through forward and backward messages. 
The system was shown able to learn backpropagation and can be used to improve upon it. 
Our proposed method OP-LSTM can also learn to implement backpropagation (see Section~\ref{sec:relation}). 
Other works \citep{ravi2017optimization, andrychowicz2016learning} have also used an LSTM for meta-learning the weight update procedure. 
Instead of feeding the training examples into the LSTM, as done by the plain LSTM \citep{hochreiter2001learning, younger2001meta}, the LSTM was fed gradients so that it could propose weight updates for a separate base-learner network.  
Our proposed method OP-LSTM is similar to these two approaches that meta-learn the weight update rules as we use an LSTM to update the weights (2D hidden states) of a base-learner network.
Note that this strategy thus also deviates from the plain LSTM approach, which is fed raw input data. 
In our approach, the LSTM acts on predictions and ground-truth targets or messages. 
In addition, we use a coordinate-wise architecture where the same LSTM is applied to different nodes in the network. 
A difference with other learning-to-optimize works \citep{ravi2017optimization, andrychowicz2016learning} is that we do not feed gradients into the LSTM and that we update the weights (2D hidden states) through outer product update rules.

\section{Meta-learning with LSTM}

In this section, we briefly review the LSTM architecture \citep{hochreiter1997long}, explain the idea of meta-learning with an LSTM through backpropagation as proposed by \citet{hochreiter2001learning} and \citet{younger2001meta}, discuss two problems with this approach in the context of meta-learning and propose solutions to solve them.

Additionally, we propose solutions to this problem. 
We prove that a single-layer RNN followed by a linear layer is incapable of embedding a classification learning algorithm in its recurrent dynamics and show by example that an LSTM adding a single linear layer is sufficient to achieve this type of learning behavior in a simple setting.

\subsection{LSTM architecture}

\begin{figure}
    \centering
    \includegraphics[scale=0.4]{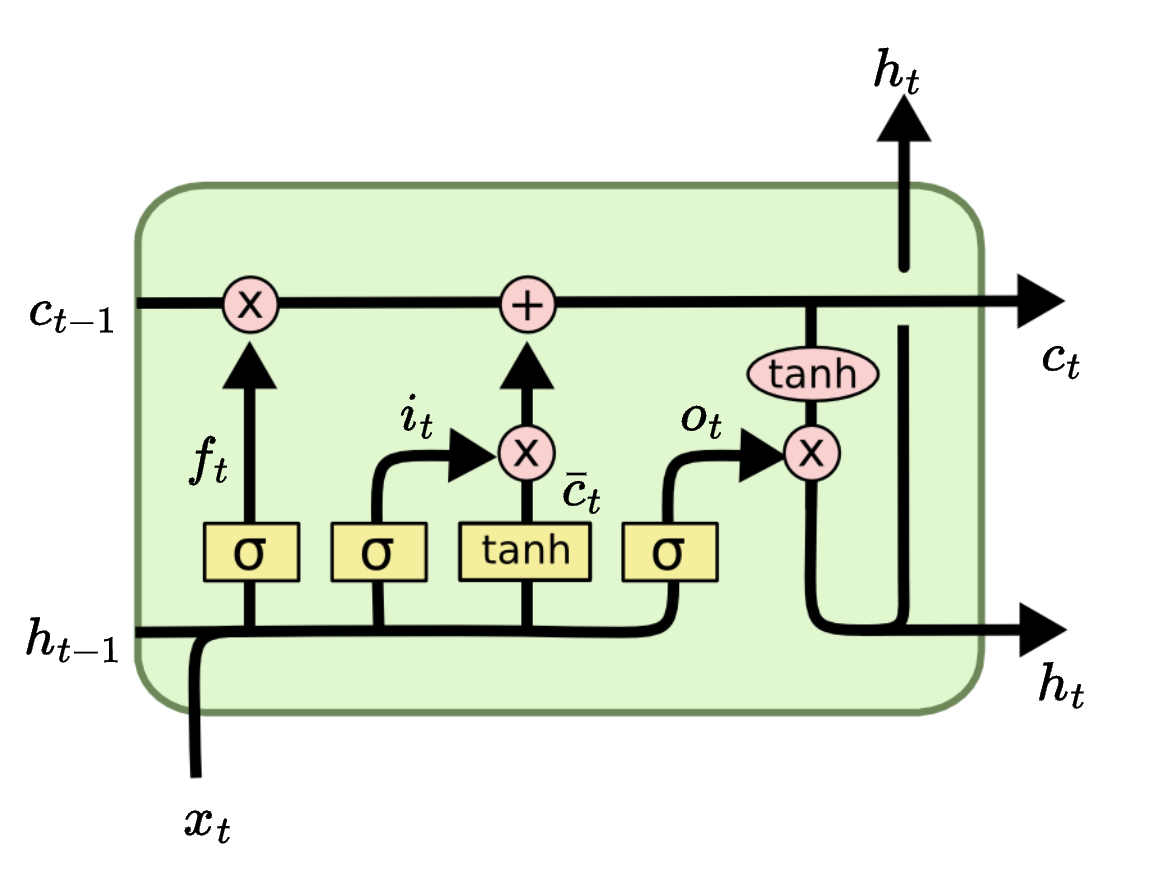}
    \caption{The architecture of an LSTM cell. The LSTM maintains an inner cell state $\mathbf{c}_t$ and hidden state $\mathbf{h}_t$ over time that are updated with new incoming data $\mathbf{x}_t$. The forget $\fgate_t$, input $\igate_t$, output $\ogate_t$, and cell $\cgate_t$ gates regulate how these states are updated. Image adapted from \citet{ColahLSTM}.}
    \label{fig:lstm}
\end{figure}

LSTM \citep{hochreiter1997long} is a recurrent neural network architecture suitable for processing sequences of data.
The architecture of an LSTM cell is displayed in Figure~\ref{fig:lstm}. 
It maintains an internal state and uses four gates to regulate the information flow within the network
\begin{align}
    \fgate_t &= \sigma ( \weight_f [\hid_{t-1}, \inp_t] + \bias_f), \\
    \igate_t &= \sigma ( \weight_i [\hid_{t-1}, \inp_t] + \bias_i), \\
    \ogate_t &= \sigma ( \weight_o [\hid_{t-1}, \inp_t] + \bias_o), \\
    \cgate_t &= \tanh ( \weight_c [\hid_{t-1}, \inp_t] + \bias_c).
\end{align}
Here, $\theta =\{\weight_f, \weight_c, \weight_i, \weight_o, \bias_f, \bias_c, \bias_i, \bias_o  \}$ are the parameters of the LSTM, $[\mathbf{a}, \mathbf{b}]$ represents the concatenation of $\mathbf{a}$ and $\mathbf{b}$, $\sigma$ is the sigmoid function (applied element-wise) and $\fgate_t, \igate_t, \ogate_t,\cgate_t \in \mathbb{R}^{d_h}$ are the forget, input, output, and cell gates, respectively. 
These gates regulate the information flow within the network to produce the next cell and hidden states
\begin{align}
    \cell_t &= \fgate_t \odot \cell_{t-1} + \igate_t \odot \cgate_t, \label{eq:cellstate}\\
    \hid_t &= \ogate_t \odot \tanh ( \cell_t ).
\end{align}

The hidden state and cell state are obtained by applying LSTM $g_{\theta}$ to inputs $(\inp_t, \inp_{t-1}, \ldots, \inp_1)$, i.e.,
\begin{align}
    [\hid_t, \cell_t] &= g_{\theta}(\inp_t, \inp_{t-1}, \ldots, \inp_1) \\
    &= m_{\theta}(\inp_t; \hid_{t-1}, \cell_{t-1}).
\end{align}

\subsection{Meta-learning with LSTM}
\citet{hochreiter2001learning} and \citet{younger2001meta} show that the LSTM can perform learning purely through unrolling its hidden state over time with fixed weights. 
When presented with a new task $\Tau_j$---denoting the concatenation of an input and its target as $\inp_t' = (\inp_t, \out_t)$---the support set $\Dtr =\{ (\inp_1,\out_1), \ldots, (\inp_M, \out_M)\} = \{ \inp_1', \inp_2', \ldots, \inp_M' \}$, is fed as a sequence, e.g., $(\inp_1, \mathit{null}), (\inp_2, \out_1), \ldots, (\inp_M, \out_{M-1})$, into the LSTM to produce a hidden state $\hid_{M}(\Dtr)$.
Predictions for unseen inputs (queries) $\hat{\inp}$ are then conditioned on the hidden state $\hid_M(\Dtr)$ and cell state $\cell_M(\Dtr)$, where we have made it explicit that $\hid_M$ and $\cell_M$ are functions of the support data.
More specifically, the hidden state of the query input $\hat{\inp} = [\inp, \out_M]$ is computed as $[\hat{\hid}, \hat{\cell}] = m_{\theta}(\hat{\inp}; \hid_M(\Dtr), \cell_M(\Dtr))$, and this hidden state is used either directly for prediction or can be fed into a classifier function (which also uses fixed weights).
Since the weights of the LSTM are fixed when presented with a new task, the learning takes place in the recurrent dynamics, and the hidden state $\hid_M(\Dtr)$ is responsible for guiding predictions on unseen inputs $\hat{\inp}$.
Note that there are different ways to feed the support data into the LSTM, as one can also   use additional data such as the error on the previous input or feed the current input together with its target tuples $(\inp_t, \out_t)$ (as done in Figure~\ref{fig:seqprocessing} and our implementation).
We use the latter strategy in our experiments as we found it to be most effective.

This recurrent learning algorithm can be obtained by performing meta-training on various tasks which require the LSTM to perform learning through its recurrent dynamics.
Given a task, we feed the training data into the LSTM, and then feed in query inputs to make predictions. 
The loss on these query predictions can be backpropagated through the LSTM to update the weights across different tasks. 
Note, however, that during the unrolling of the LSTM over the training data, the weights of the LSTM are held fixed.  
The weights are thus only updated across different tasks (not during adaptation to individual tasks) to improve the recurrent learning algorithm. 
By adjusting the weights of the LSTM using backpropagation across different tasks, we are essentially changing the learning program of the LSTM and hence performing meta-learning.

\subsection{Problems with the classical LSTM architecture}
The classical LSTM architecture suffers from two issues that may limit its ability to implement recurrent learning algorithms.

\paragraph{Non-temporal training data} LSTMs work with sequences of data. When using an LSTM in the meta-learning context, the recurrent dynamics should implement a learning algorithm and process the support dataset.
This support dataset $\Dtr = \{ (\inp_1, \out_1), \ldots, (\inp_M, \out_M) \} = \{ \inp_1', \inp_2', \ldots, \inp_M' \}$, however, is a \emph{set} rather than a sequence.
This means that we would want the hidden embedding after processing the support data to be invariant with respect to the order in which the examples are fed into the LSTM. 
Put more precisely, given any two permutations of the $M$ training examples $\pi = (\pi_1, \pi_2, \ldots, \pi_M)$ and $\pi' = (\pi_1', \pi_2', \ldots, \pi_M')$, we want to enforce
\begin{align}
    g_{\theta}(\inp_{\pi_1}', \inp_{\pi_2}', \ldots, \inp_{\pi_M\vphantom{'}}') = g_{\theta}(\inp_{\pi_1'}', \inp_{\pi_2'}', \ldots, \inp_{\pi_M'}'),
\end{align} where $\inp_{\pi_i}'$ is the $i$-th input (possibly containing target or error information) under permutation $\pi$ and $\inp_{\pi_i'}'$ the input under permutation $\pi'$.

\paragraph{Intertwinement of embedding and learning}
In the LSTM approach proposed by \citet{hochreiter2001learning} and \citet{younger2001meta}, the recurrent dynamics implement a learning algorithm.
At the same time, however, the hidden state also serves as an input embedding.
Thus, in this approach, the input embedding and learning procedures are intertwined. 
This may be problematic because a learning procedure may be highly complex and nonlinear, whilst the optimal input embedding may be simple and linear. 
For example, suppose that we feed convolutional features into a plain LSTM.
Normally, we often compute predictions using a linear output layer. 
Thus, a simple single-layer LSTM may be the best in terms of input representation. 
However, the learning ability of a single-layer LSTM may be too limited, leading to bad performance. 
In other words, stacking multiple LSTM layers may be beneficial for finding a better learning algorithm, but the resulting input embedding may be too complex, which can lead to overfitting. On the other hand, a good but simple input embedding may overly restrict the search space of learning algorithms, resulting in a bad learning algorithm.  

An LSTM with sufficiently large hidden dimensionality may be able to separate the learning from the input representation by using the first $N$ dimensions of the hidden representations to perform learning and to preserve important information for the next time step, and using the remaining dimensions to represent the input. 
However, this poses a challenging optimization problem due to the risk of overfitting and the large number of parameters that would be needed.

\subsection{Towards an improved architecture}\label{sec:oplstm}

These potential issues of the classical LSTM architecture inspire us to develop an architecture that is better suited for meta-learning. 

\begin{figure}
    \centering
    \includegraphics[width=\textwidth]{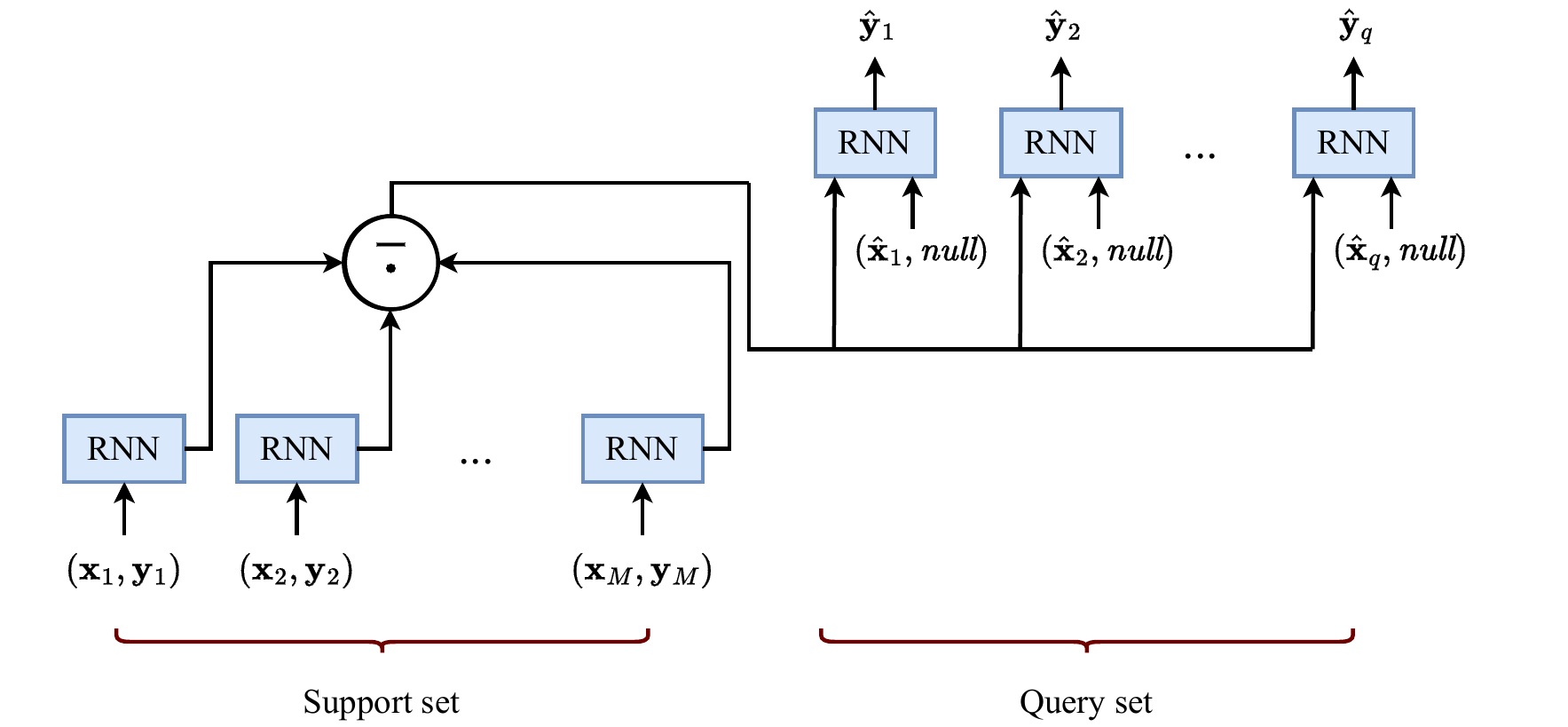}
    \caption{Our proposed batch processing of the support data, resulting in a state that is permutation invariant. Every support example $(\inp_i, \out_i)$ is processed in parallel, and the resulting hidden states are aggregated with mean-pooling (denoted by the symbol $\bar{\cdot}$). The predictions $\hat{\out}_j$ for new query points $\hat{x}_j$ are conditioned on the resulting permutation-invariant state. Note that the support data is only fed once into the RNN (a single time step $t$), although it is possible to make multiple passes over the data, by feeding the mean-pooled state into the RNN at the next time step.}  
    \label{fig:batchprocessing}
\end{figure}

\paragraph{Non-temporal data $\rightarrow$ average pooling}
In order to enforce invariance of the hidden state and cell state with respect to the order of the support data, we can \emph{pool} the individual embeddings. 
That is, given an initial state of the LSTM  $\mathbf{s}_t = [\hid_t, \cell_t]$, we update the state by processing the support data as a batch and by average pooling, i.e., 
\begin{align}
    \mathbf{s}_{t+1} = \left[\hid_{t+1}, \cell_{t+1} \right] = \frac{1}{M}\sum_{i=1}^{M} m_{\theta}(\inp_i'; \hid_{t}, \cell_t). \label{eq:pooling}
\end{align}
Note that one time step now corresponds to processing the entire support dataset once, since $\mathbf{s}_{t+1}$ is a function thereof. 
Our proposed batch processing for a single time step (during which we ingest the support data) has been visualized in Figure~\ref{fig:batchprocessing}.

\paragraph{Intertwinement of embedding and learning}
The problems associated with the intertwinement of the embedding and learning procedures can be solved by decoupling them. 
In this way, we create two separate procedures: 1)~the embedding procedure, and 2)~the learning procedure. 
The embedding procedure can be implemented by a base-learner neural network, and the learning procedure by a meta-network that updates the weights of the base-learner network.

In the plain LSTM approach, where the learning procedure is intertwined with the input representation mechanism, predictions would be conditioned on the hidden state $\hid(\inp, \hid(\Dtr), \cell(\Dtr))$.
Instead, we choose to use the inner product between the hidden state (acting as weight vector) and the embedding of current input $\act^{(L)}(\inp)$, i.e., 
\begin{align}
    \hat{y}(\inp) = a^{(L)}(\inp) = \underbrace{\hid^{(L)}(\Dtr)^T}_{\text{learning}} \underbrace{\act^{(L-1)}(\inp) \vphantom{\hid(\Dtr)^T} }_{\text{embedding}},
\end{align} where $\act^{(L-1)}(\inp)$ is the representation of input $\inp$ in layer $L - 1$ of some base-learner network (consisting of $L$ layers), whose weights are updated by a meta-network.
We use the inner product to force interactions between the learning and embedding components, so that the predictions can not rely on either of the two separately.
Note that by computing predictions in this way, we effectively decouple the learning algorithm implemented by hidden state dynamics from the input representation. 
A problem with this approach is that the output is a single scalar. 
In order to obtain an arbitrary output dimension $d_{out} > 1$, we should multiply the input representation $\act^{(L-1)}(\inp)$ with a matrix $\mathbf{H} \in \mathbb{R}^{d_{out} \times d_{in}}$, i.e., $\mathbf{\hat{y}}(\inp) = \mathbf{H}^{(L)} \act^{(L-1)}(\inp)$.
In order to obtain $\mathbf{H}^{(L)}$, one could use a separate LSTM with a hidden dimension of $d_{in}$ per output dimension, but the number of required LSTMs would grow linearly with the output dimensionality. 
Instead, we use the outer product, which requires only one hidden vector of size $d_{in}$ that can be outer-multiplied with a vector of size $d_{out}$.  
We detail the computation of 2D weight matrices $\mathbf{H}$ (hidden states) with the outer product rule in the next section. 

Note that the 2D hidden state $\mathbf{H}$ can be seen as a weight matrix of a regular feed-forward neural network, which allows us to generalize this approach to networks with an arbitrary number of layers, where we have a 2D hidden state $\mathbf{H}^{(\ell)}$ for every layer $\ell \in \{ 1,2,\ldots,L \}$ in a network with $L$ layers.
Our approach can then be seen as meta-learning an outer product weight update rule for the base-learner network such that it can quickly adapt to new tasks.

\section{Outer product LSTM (OP-LSTM)}

Here, we propose a new technique, called \emph{Outer Product LSTM (OP-LSTM)}, based on the problems of the classical LSTM architecture for meta-learning and our suggested solutions.
We begin by discussing the architecture, then cover the learning objective and algorithm, and end by studying the relationship between OP-LSTM and other methods.  

\subsection{The architecture}
Since we can view the 2D hidden states $\mathbf{H}^{(\ell)}$ in OP-LSTM as weight matrices that act on the input, the OP-LSTM can be interpreted as a regular fully-connected neural network.
The output of the OP-LSTM for a given input $\inp$ is given by \begin{align}
    f_{\theta}(\inp, \Dtr, T) = \sigma^{(L)}(\mathbf{H}_T^{(L)} \act^{(L-1)}_T(\inp) + \bias^{(L)}),
\end{align} where $\Dtr$ is the support dataset of the task, $L$ the number of layers of the base-learner network, and $T$ the number of time steps that the network unrolls (trains) over the entire support set. 
Here, $\sigma^{(L)}$ is the activation function used in layer $L$, $\bias^{(L)}$ the bias vector in the output layer, and $\act^{(L-1)}_T(\inp)$ the input to layer $L$ after making $T$ passes over the support set and having received the query input.

\begin{figure}
    \centering
    \includegraphics[width=\textwidth]{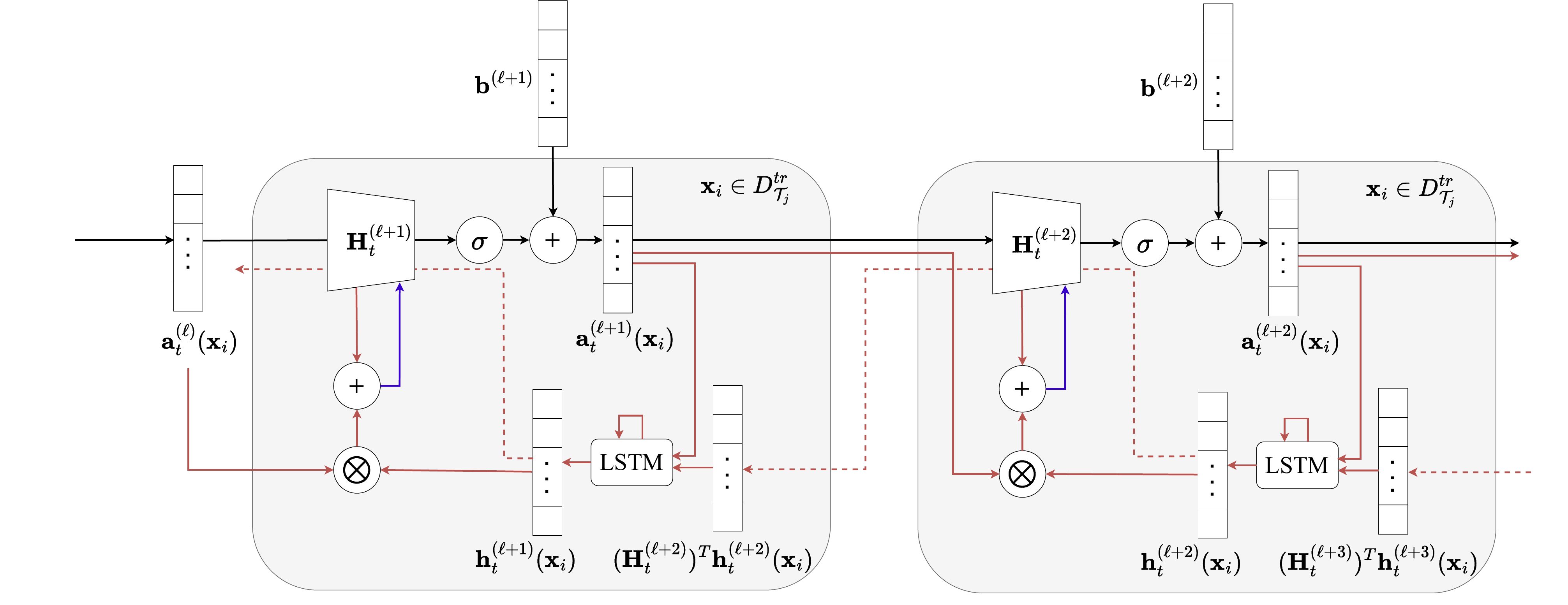}
    \caption{The workflow of OP-LSTM. We have visualized two layers of the base-learner network. During the forward pass, the 2D hidden states $\mathbf{H}^{(\ell)}_t$ act as weight matrices of a feed-forward neural network that act on the input of that layer $\act^{(\ell - 1)}_t$. This linear combination $\mathbf{H}^{(\ell)}_t \act^{(\ell - 1)}_t(\inp_i)$ is passed through a nonlinearity $\sigma$ and added with a bias vector $\mathbf{b}^{(\ell)}$ to produce the activation $\act^{(\ell)}_t(\inp_i)$.  The entire forward pass is displayed by the black arrows. The red arrows, on the other hand, indicate the backward pass using the coordinate-wise LSTM. The outer product ($\otimes$) of the resulting hidden state $\hid_{t}^{(\ell+1)}$ and the inputs from the previous layer $\act^{(\ell)}_t$ are added to the 2D hidden state $\mathbf{H}_t^{(\ell + 1)}$ to produce $\mathbf{H}_{t+1}^{(\ell + 1)}$ (blue arrow), which can be interpreted as the updated weight matrix.}
    \label{fig:oplstm}
\end{figure}

Put more precisely, the activation in layer $\ell$ at time step $t$, as a function of an input $\mathbf{x}$, is denoted $\act^{(\ell)}_t(\inp)$ and  defined as follows
\begin{align}
    \act^{(\ell)}_t(\inp) = \begin{cases}
    \inp & \text{if $\ell=0$ (input layer), }\\
    \sigma^{(\ell)}(\mathbf{H}^{(\ell)}_{t} \act^{(\ell - 1)}_t(\inp) + \bias^{(\ell)} ) & \text{otherwise.}
    \end{cases} \label{eq:forwardpass}
\end{align} Note that this defines the forward dynamics of the architecture.  
Here, the $\mathbf{H}^{\ell}_t \in \mathbb{R}^{d_{out}^{(\ell)} \times d_{in}^{(\ell)}}$ is the 2D hidden state that is updated by pooling over the normalized 2D \emph{outer product} hidden states $\hid_{t+1}^{(\ell)}(\inp'_i) \act_{t}^{(\ell - 1)}(\inp_i')^T$  associated with individual training examples $\inp'_i = (\inp_i, \out_i)$, i.e.,   
\begin{align}
    \mathbf{H}^{(\ell)}_{t+1} = \mathbf{H}^{(\ell)}_{t} + \frac{\gamma}{M} \sum_{i=1}^{M} \frac{\hid_{t+1}^{(\ell)}(\inp'_i) \act_{t}^{(\ell - 1)}(\inp_i)^T}{|| \hid_{t+1}^{(\ell)}(\inp'_i) \act_{t}^{(\ell - 1)}(\inp_i)^T ||_F}, \label{eq:updateH}
\end{align} where $\gamma$ is the step size of the updates and $|| \cdot ||_F$ is the Frobenius norm.
We perform this normalization for numerical stability. 
Note that this update using average pooling ensures that the resulting hidden states $\mathbf{H}^{(\ell)}_t$ are invariant to permutations of the support data.  
Moreover, we observe that this equation defines the backward dynamics of the architecture (updating the 2D hidden states).
However, this equation does not yet tell us how the hidden states $\hid^{(\ell)}_{t+1}(\inp'_i)$ are computed.

We use a coordinate-wise LSTM so that the same LSTM can be used in layers of arbitrary dimensions, in similar fashion as \citet{ravi2017optimization, andrychowicz2016learning}.
This means that we maintain a state $s_{t,j}^{(\ell)} = [h^{(\ell)}_{t,j}, c^{(\ell)}_{t,j}]$ for every individual node $j$ in the state vector and every layer $\ell \in \{1,2,\ldots,L\}$ over time steps $t$. 
In order to obtain the hidden state vector for a given layer $\ell$ and time step $t$, we simply concatenate the individual hidden states computed by the coordinate-wise LSTM, i.e., $\hid_t^{(\ell)} = [h^{(\ell)}_{t,1}, h^{(\ell)}_{t,2}, \ldots, h^{(\ell)}_{t,d^{(\ell)}}]^T$, where $d^{(\ell)}$ is the number of neurons in layer $\ell$. 
The LSTM weights to update these states are shared across all layers and nodes with the same activation function. For classification experiments, we often have two LSTMs: one for the final layer which uses a softmax activation function, and one for the body of the network, which uses the ReLU activation. 
This allows OP-LSTM to learn weight updates akin to gradient descent, where the backward computation is tied to each nonlinearity in the base-learner network (as this can not be done by a single LSTM, due to the different non-linearities of the softmax and the RELU activation). 
We use pooling over the support data in order to update the states using a coordinate-wise approach, where every element of the hidden state $\hid^{(\ell)}_t$ of a given layer $\ell$ is updated independently by a single LSTM.

In order to compute the next state $s^{(\ell)}_{t+1,j}$ of node $j$ in layer $\ell$, we need to have the previous state consisting of the previous hidden state $h_{t,j}^{(\ell)}$ and cell state $c_{t,j}^{(\ell)}$ of that node.
Moreover, we need to feed the LSTM an input, which we define as $z_{t,j}^{(\ell)}$.
In OP-LSTM, we define this input as
\begin{align}
    z_{t,j}^{(\ell)}(\inp_i') = \begin{cases} 
    [(\act_t^{(\ell)}(\inp_i))_j, (\out_i)_j] & \text{if $\ell = L$ (output layer),} \\
    [(\act_t^{(\ell)}(\inp_i))_j, ((\mathbf{H}^{(\ell + 1)}_t)^T  \hid^{(\ell+1)}_t(\inp_i))_j] & \text{otherwise}. \label{eq:message}
    \end{cases}
\end{align}
Note that for the output layer $L$, the input to the LSTM corresponds to the current prediction and the ground-truth output, which share the same dimensionality.
For the earlier layers in the network, we do not have access to the ground-truth signals.
Instead, we view the hidden state of the output layer LSTM as errors and propagate them backward through the 2D hidden states $\mathbf{H}^{(\ell + 1)}_t$, hence the expression $(\mathbf{H}^{(\ell + 1)}_t)^T  \hid^{(\ell+1)}_t(\inp_i)$ for earlier layers. 
We note that this is akin to backpropagation, where error messages $\delta^{(\ell + 1)}$ are passed backward through the weights of the network.

Given an input $\inp_i'$, the next state $s^{(\ell)}_{t+1,j}$ can then be computed by applying the LSTM $m_{\theta}$ to the input $z_{t,j}^{(\ell)}(\inp_i')$, conditioned on the previous hidden state $h_{t,j}^{(\ell)}$ and cell state $c_{t,j}^{(\ell)}$. 
\begin{align}
    s^{(\ell)}_{t+1,j}(\inp_i') = [h^{(\ell)}_{t+1,j}(\inp_i'),c^{(\ell)}_{t+1,j}(\inp_i')] = m_{\theta}(z_{t,j}^{(\ell)}(\inp_i'); h_{t,j}^{(\ell)},c_{t,j}^{(\ell)}), \label{eq:indivstate}
\end{align} 
where $z_{t,j}^{(\ell)}(\inp_i')$ is the input to the LSTM used to update the state. 
These individual states are averaged over all training inputs to obtain
\begin{align}
    s^{(\ell)}_{t+1,j} = [h^{(\ell)}_{t+1,j},c^{(\ell)}_{t+1,j}] = \frac{1}{M}\sum_{i=1}^{M}\label{eq:avgstate} s^{(\ell)}_{t+1,j}(\inp_i').
\end{align}
Note that we can obtain a state vector, hidden vector, and cell state vector, by concatenation, i.e., $\mathbf{s}^{(\ell)}_{t+1,j}(\inp_i') \allowbreak = [s^{(\ell)}_{t+1,1}(\inp_i'), \allowbreak s^{(\ell)}_{t+1,2}(\inp_i'), \allowbreak \ldots, \allowbreak s^{(\ell)}_{t+1,d^{(\ell)}}(\inp_i')]$, $\mathbf{h}^{(\ell)}_{t+1,j}(\inp_i') \allowbreak = [h^{(\ell)}_{t+1,1}(\inp_i'), \allowbreak h^{(\ell)}_{t+1,2}(\inp_i'), \allowbreak \ldots, \allowbreak h^{(\ell)}_{t+1,d^{(\ell)}}(\inp_i')]$, and $\mathbf{c}^{(\ell)}_{t+1,j}(\inp_i') \allowbreak = [c^{(\ell)}_{t+1,1}(\inp_i'), \allowbreak c^{(\ell)}_{t+1,2}(\inp_i'), \allowbreak \ldots, \allowbreak c^{(\ell)}_{t+1,d^{(\ell)}}(\inp_i')]$.

\subsection{The algorithm}

OP-LSTM is trained to minimize the expected loss on the query sets conditioned on the support sets, where the expectation is with respect to a distribution of tasks.
Put more precisely, we wish to minimize $\expect_{\Tau_j \backsim p(\Tau)} \left[  \loss_{\Dte}(\Theta)  \right]$, where $\Theta = \{ \mathbf{\theta}, \mathbf{H}^{(1)}_0,  \mathbf{H}^{(2)}_0, \ldots,  \mathbf{H}^{(L)}_0, \bias^{(1)}, \bias^{(2)}, \ldots, \bias^{(L)} \}$.
This objective can be approximated by sampling batches of tasks, updating the weights using the learned outer product rules, and evaluating the loss on the query sets.
Across tasks, we update $\Theta$ using gradient descent. 
In practice, we use the cross-entropy loss for classification tasks and the MSE loss for regression tasks.

The pseudocode for OP-LSTM is displayed in Algorithm~\ref{alg:oplstm}. 
First, we randomly initialize the initial 2D hidden states $\mathbf{H}^{(\ell)}_0$ and the LSTM parameters $\mathbf{\theta}$.
We group these parameters as $\Theta = \{ \mathbf{\theta}, \mathbf{H}^{(1)}_0,  \mathbf{H}^{(2)}_0, \ldots,  \mathbf{H}^{(L)}_0  \} $, which will be meta-learned across different tasks.
Given a task $\Tau_j$, we make $T$ updates on the entire support set $\Dtr$ by processing the examples individually, updating the 2D hidden states $\mathbf{H}^{(\ell)}_t$, and computing the new hidden states of the coordinate-wise LSTM for every layer $\mathbf{s}^{(\ell)}_t$. 
After having made $T$ updates on the support data, we compute the loss of the model on the query set $\Dte$.
The gradient of this loss with respect to all parameters $\Theta$ is added to the gradient buffer.
Once a batch of tasks $B$ has been processed in this way, we perform a gradient update on $\Theta$ and repeat this process until convergence or a maximum number of iterations has been reached.

\begin{algorithm}
\caption{Meta-learning with outer product LSTM (OP-LSTM)}\label{alg:oplstm}
\begin{algorithmic}[1]
\STATE Randomly initialize $\mathbf{H}^{(\ell)}_0$ and biases $\bias^{(\ell)}$ for all $1 \leq \ell \leq L$ 
\STATE Randomly initialize LSTM parameters $\theta$, set $\hid_o^{(\ell)} = \mathbf{0}$, $\cell_0^{(\ell)} = \mathbf{0}$
   \REPEAT
        \STATE Initialize gradient buffer $\zeta = \mathbf{0}$
        \STATE Sample batch of J tasks $B = \{ \Tau_j \backsim p(\Tau) \}_{j=1}^J$
        \FOR{$\Tau_j = (D^{tr}_{\Tau_j}, D^{te}_{\Tau_j})$ in $B$}
            \STATE set $\hid_o^{(\ell)} = \mathbf{0}$, $\cell_0^{(\ell)} = \mathbf{0}$ for all $1 \leq \ell \leq L$ 
            \FOR{$t = 1,...,T$}
                \FOR{$\inp_i' = (\inp_i, \out_i) \in \Dtr$}
                    \STATE Compute predictions $\act^{(L)}_{t-1}(\inp_i)$ (Eq.~\ref{eq:forwardpass})
                    \STATE Compute $\mathbf{z}^{(\ell)}_{t}(\inp_i')$ and $\hid^{(\ell)}_{t}(\inp_i)$ for $1 \leq \ell \leq L$ with backward message passing (see Eq.~\ref{eq:message} and Eq.~\ref{eq:indivstate})
                    \STATE Update $\mathbf{H}^{(\ell)}_{t}$ for $1 \leq \ell \leq L$  (Eq.~\ref{eq:updateH})
                \ENDFOR
                \STATE Compute $\mathbf{s}^{(\ell)}_{t}$ for $1 \leq \ell \leq L$ (Eq.~\ref{eq:avgstate}) through concatenation
            \ENDFOR
            \STATE Compute query predictions $\loss_{\Dte}(\{ \mathbf{H}^{(\ell)}_T \}_{\ell=1}^L)$
            \STATE Update gradient buffer $\zeta = \zeta + \frac{1}{J} \nabla_{\Theta}\loss_{\Dte}(\{ \mathbf{H}^{(\ell)}_T \}_{\ell=1}^L)$
        \ENDFOR
        \STATE Update $\boldsymbol{\Theta} = \Theta - \beta \zeta$ 
   \UNTIL{convergence}
\end{algorithmic}
\end{algorithm}

\section{Experiments}

In this section, we aim to answer the following research questions:
\begin{itemize}
\item How do the performance and training stability of a plain LSTM compare when processing the support data as a sequence versus as a set with average pooling? (see Section~\ref{sec:perminvariance})
\item How well does the plain LSTM perform at few-shot sine wave regression and within- and cross-domain image classification problems compared with popular meta-learning methods such as MAML \citep{finn2017model} and Prototypical network \citep{snell2017prototypical}? (see Section~\ref{sec:sinewave} and Section~\ref{sec:images})
\item Does OP-LSTM yield a performance improvement over the simple LSTM and the related approaches MAML and Prototypical network in few-shot sine wave regression and within- and cross-domain image classification problems? (see Section~\ref{sec:sinewave} and Section~\ref{sec:images})
\item How does OP-LSTM adjust the weights of the base-learner network? (see Section~\ref{sec:howupdate})
\end{itemize}

For our experiments, we use few-shot sine wave regression \citep{finn2017model} as an illustrative task, and popular few-shot image classification benchmarks, namely Omniglot \citep{lake2015human}, miniImageNet \citep{ravi2017optimization, vinyals2016matching}, and CUB \citep{wah2011caltech}. 
We use MAML \citep{finn2017model}, prototypical network \citep{snell2017prototypical}, SAP \citep{huisman2023} and Warp-MAML \citep{Flennerhag2020Meta-Learning} as baselines. 
The former two  are popular meta-learning methods and can both be approximated by the OP-LSTM (see Section~\ref{sec:relation}), allowing us to investigate the benefit of OP-LSTM's expressive power.
The last two baselines are used to investigate how OP-LSTM compares to state-of-the-art gradient-based meta-learning methods in terms of performance, although it has to be noted that that OP-LSTM is orthogonal to that method, in the sense that OP-LSTM could be used on top of Warp-MAML.
However, this is a nontrivial extension and we leave this for future work. 
We run every technique on a single GPU (PNY GeForce RTX 2080TI) with a computation budget of 2 days (for detailed running times, please see Section~\ref{sec:runningtimes}). 
Each experiment is performed with 3 different random seeds, where the random seed affects the random weight initialization of the neural networks as well as the used training tasks, validation tasks, and testing tasks. 
Below, we describe the different experimental settings that we use. 
Note that we do not aim to achieve state-of-the-art performance, but rather investigate whether the plain LSTM is a competitive method for few-shot learning on modern benchmarks and whether OP-LSTM yields improvements over the plain LSTM, MAML, and Prototypical network.   

\paragraph{Sine wave regression}
This toy problem was originally proposed by \citet{finn2017model} to study meta-learning methods. 
In this setting, every task $\Tau_j$ corresponds to a sine wave $s_j = A_j \cdot \sin(x - p_j)$, where $A_j$ and $p_j$ are the amplitude and phase of the task, sampled uniformly at random from the intervals [0.1, 5.0] and [0, $\pi$], respectively.
The goal is to predict for a given task the correct output $y$ given an input $x$ after training on the support set, consisting of $k$ examples.
The performance of learning is measured in the query set, consisting of 50 input-output. 
For the plain LSTM approach, we use a multi-layer LSTM trained with Backpropagation through Time (BPTT) using Adam \citep{kingma2015adam}.
During meta-training, the LSTM is shown 70\,000 training tasks.
Every 2\,500 tasks, we perform meta-validation on 1\,000 tasks, and after having selected the best validated model, we evaluate the performance on 2\,000 meta-test tasks.

\paragraph{Few-shot image classification}

In case of few-shot image classification, all methods are trained for 80\,000 episodes on training tasks and we perform meta-validation every 2\,500 episodes.
The best learner is then evaluated on 600 hold-out test tasks, each task having a number of examples per class in the support set as indicated by the experiment (ranging from 1--10) as well as a query set of 15 examples per class.
We repeat every experiment 3 times with different random seeds, meaning the that weight initializations and tasks are different across runs, although the class splits for sampling training/validation/testing tasks are kept fixed.
For the Omniglot image classification dataset, we used a fully-connected neural network as base-learner for MAML and OP-LSTM, following \citet{Santoro16} and \citet{finn2017model}. 
The network consists of 4 fully-connected blocks with dimensions 256-128-64-64. Every block consists of a linear layer, followed by BatchNorm and ReLU activation. 
Every layer of the base-learner network is an OP-LSTM block. 
The plain LSTM approach uses an LSTM as base-learner. 
For MAML, we use the best reported hyperparameters by \citet{finn2017model}.
We performed hyperparameter tuning for LSTM and OP-LSTM using random search and grid search, respectively (details can be found in Appendix~\ref{sec:hypers}). 
Note that as such, the comparison against MAML and Prototypical networks is only for illustrative purposes, as the hyperparameter optimization procedure on these methods has, due to computational restrictions, not been executed under the same conditions.

For the miniImageNet and CUB image classification datasets, we use the Conv-4 base-learner network for all methods, following \citet{snell2017prototypical, finn2017model}.
This base-learner consists of 4 blocks, where every block consists of 64 feature maps created with $3 \times 3$ kernels, BatchNorm, and ReLU nonlinearity. 
MAML uses a linear output layer to compute predictions, the plain LSTM operates on the flattened features extracted by the convolutional layers (as an LSTM taking image data as input does not scale well), whereas OP-LSTM uses an OP-LSTM block (see Figure~\ref{fig:oplstm}) on these flattened features. 
Importantly, OP-LSTM is only used in the final layer as it does currently not support propagating messages backward through max pooling layers.

We first study the within-domain performance of the meta-learning methods, where test tasks are sampled from the same dataset as the one used for training (albeit with unseen classes).
Afterward, we also study the cross-domain performance, where the techniques train on tasks from a given dataset and are evaluated on test tasks from another dataset.
More specifically, we use the scenarios miniImageNet $\rightarrow$ CUB (train on miniImageNet and evaluate on CUB) and vice versa.

\subsection{Permutation invariance for the plain LSTM}\label{sec:perminvariance}

First, we investigate the difference in performance of the plain LSTM approach when processing the support data as a sequence $(\inp_1, \out_1), \ldots, (\inp_k, \out_k)$ or as a set $\{ ( \inp_1, \out1), \ldots, (\inp_k, \out_k) \}$ (see Section~\ref{sec:oplstm}) on few-shot sine wave regression and few-shot Omniglot classification.
For the former, every task consists of 50 query examples, whereas for the latter, we have 10 query examples per class. 
We tuned the LSTM that processes the support data sequentially with random search (details in appendix).
We compare the performance of this tuned sequential model to that of an LSTM with batching (with the same hyperparameter configuration) to see whether the resulting permutation invariance is helpful for the performance and training stability of the LSTM.
To measure the stability of the training process, we compute the confidence interval over the mean performances obtained over 3 different runs rather than over all performances concatenated for the different runs, as done in later experiments for consistency with the literature.

\begin{figure}[!htb]
    \centering
    \begin{subfigure}{0.48\textwidth}
    \includegraphics[width=\textwidth]{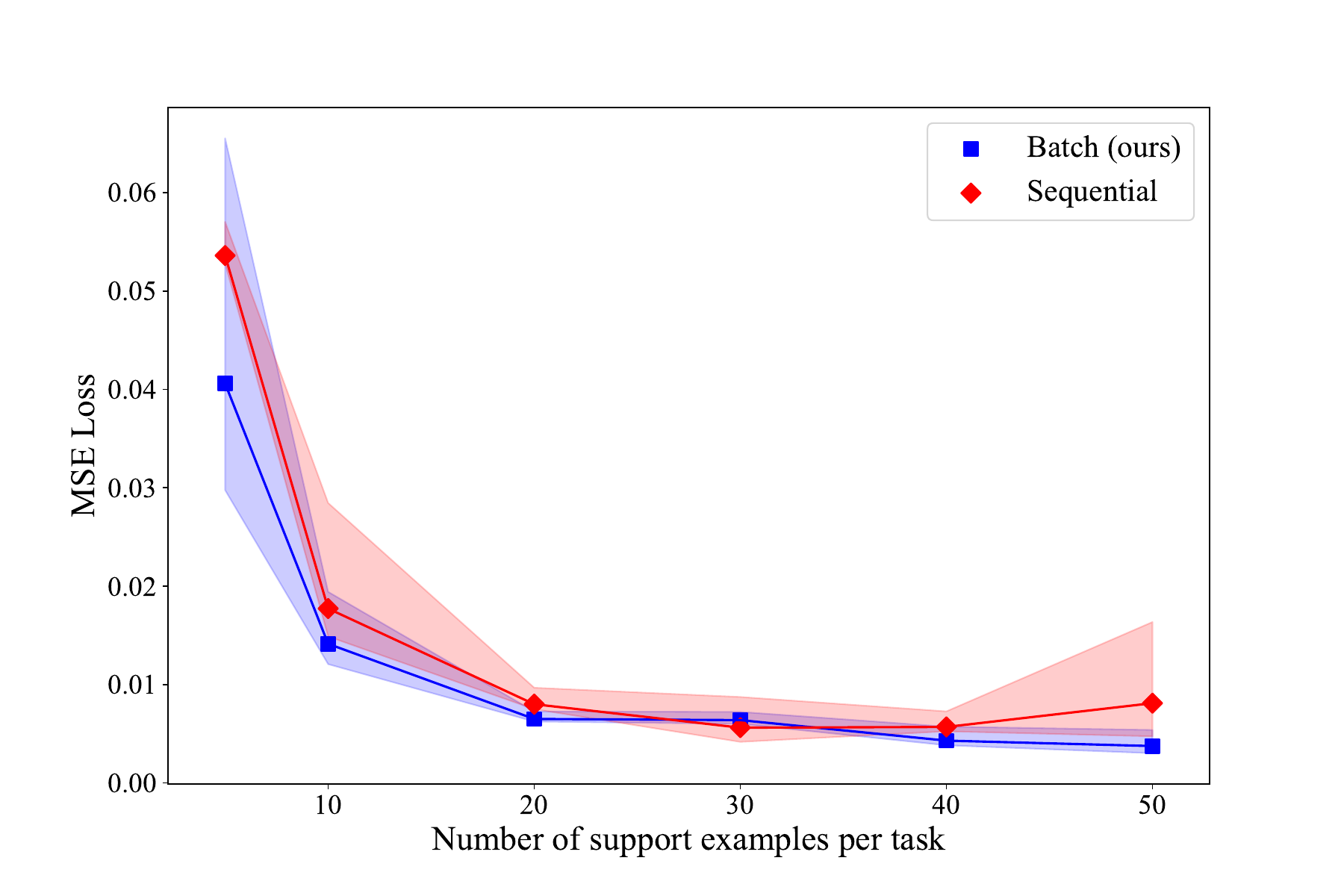}
    \caption{Sinewave - Average MSE loss}
    \end{subfigure}
    \begin{subfigure}{0.48\textwidth}
    \includegraphics[width=\textwidth]{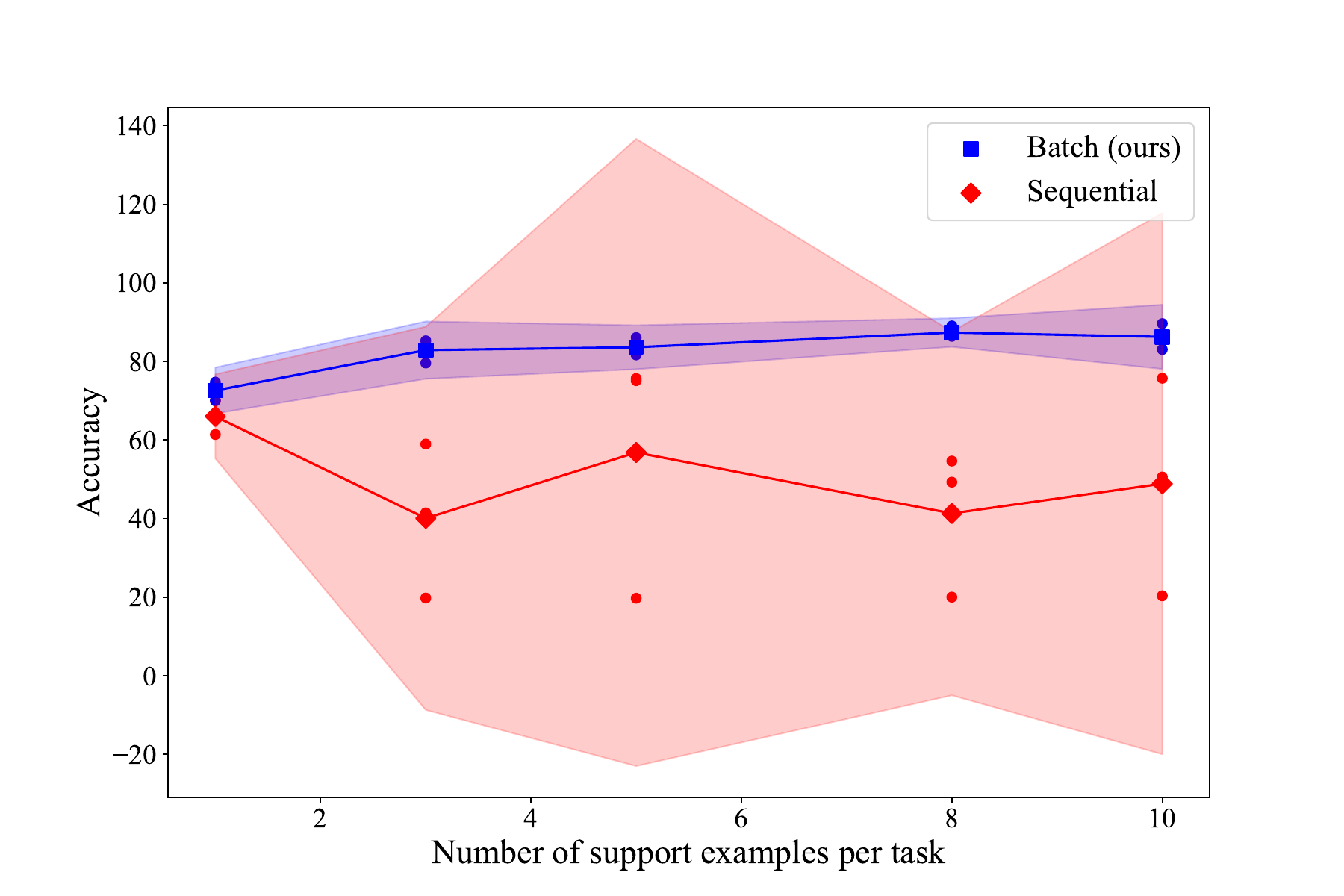}
    \caption{Omniglot - Average accuracy (\%)}
    \end{subfigure}
    \caption{The average accuracy score of a plain LSTM with sequential and batch support data processing on few-shot sine wave regression (left) and Omniglot classfication (right) for different numbers of training examples per task. Note that a lower MSE (left) or a higher accuracy (right) corresponds to better performance. The results are averaged over 3 runs (each measured over 600 meta-test tasks) with different random seeds and the 95\% confidence intervals over the mean performances of the runs are shown as shaded regions. Additionally, in the right plot, we have added scatter marks to indicate the average performances per run (dots, unconnected, 3 per setting).  Batch processing performs on par or outperforms sequential processing and improves the training stability over different runs.}
    \label{fig:supp-processing}
\end{figure}

The results of this experiment are shown in Figure~\ref{fig:supp-processing}. 
In the case of few-shot sine wave regression (left subfigure), the performance of the LSTM with batching is on par or better compared with the sequential LSTM as the MSE score of the former is smaller or equal. 
We also note that the performance tends to improve with the amount of available training data. 
A similar, although more convincing, pattern can be seen in the case of few-shot Omniglot classification (right subfigure), where the LSTM with batching significantly outperforms the sequential LSTM across the different numbers of training examples per class.
Surprisingly, in this case the performance of the LSTM does not improve as the number of examples per class increases. 
We found that this is due to training stability issues of the plain LSTM (as shown by the confidence intervals): for some runs, the LSTM does not learn and yields random performance, and in other runs the learning starts only after a certain period of burn-in iterations and fails to reach convergence within 80K meta-iterations (see appendix Section~\ref{sec:omniglot} for detailed learning curves for every run).
For the LSTM with batching, we do not observe such training stability issues. 
This shows that batching not only helps improve the performance, but also greatly increases the training stability. 
Note that the fact that the shaded confidence interval of the sequential LSTM goes above the performance obtained by the batching LSTM is an artifact of using symmetrical confidence intervals above and below the mean trend: the sequential LSTM never outperforms the batching LSTM.  
As we can see, the MSE loss for both approaches decreases as the size of the support set increases, as more training data is available for learning. 
Furthermore, we see that the performance of the LSTM with batching improves with the number of available training data, whereas this is not the case for the sequential LSTM, which struggles to yield competitive performance. 
Overall, the results imply that the permutation invariance is a helpful inductive bias to improve the few-shot learning performance.
Consequently, we will use the LSTM with batching henceforth.

\subsection{Performance comparison on few-shot sine wave regression}\label{sec:sinewave}

Next, we compare the performance of the plain LSTM with batching, our proposed OP-LSTM, as well as MAML \citep{finn2017model}. 
To ensure a fair comparison with MAML, we tuned the hyperparameters in the same way as for the plain LSTM as done in the previous subsection on 5-shot sine wave regression.
For this tuning, we used the default base-learner architecture consisting of two hidden layers with 40 ReLU nodes, followed by an output layer of 1 node. 
Afterward, we searched over different architectures with different numbers of parameters such that the expressivity in terms of the number of parameters does not limit the performance of MAML. 
We used the same base-learner architecture for the OP-LSTM as MAML without additional tuning. 

\begin{table}[!htb]
    \centering
    \caption{Average test MSE on few-shot sine wave regression. The 95\% confidence intervals are displayed as $\pm x$, and calculated over all meta-test tasks. We used batch processing for the LSTM and OP-LSTM. The best performances are dislpayed in bold font. }
    \label{tab:perfsine}
    \begin{tabular}{lrrrr}
        \toprule
        & Parameters & 5-shot & 10-shot & 20-shot \\
        \midrule
        MAML & 17\,018 & 0.18 $\pm$ 0.009 & 0.033 $\pm$ 0.003 & 0.005 $\pm$ 0.001 \\
        LSTM & 20\,201 & \textbf{0.04} $\pm$ 0.002 & 0.010 $\pm$ 0.001 & 0.007 $\pm$ 0.000\\
        OP-LSTM & 18\,107 & 0.11 $\pm$ 0.009 & \textbf{0.008} $\pm$ 0.001 & \textbf{0.003} $\pm$ 0.000 \\
        \bottomrule
    \end{tabular}
\end{table}

The test performances on the sine wave regression taska are displayed in Table~\ref{tab:perfsine}.
We note MAML, despite having a comparable number of parameters (models with more parameters than LSTM and OP-LSTM performed worse), is outperformed by LSTM and OP-LSTM, indicating that LSTM and OP-LSTM have discovered more efficient learning algorithms for sine wave tasks. 
Comparing LSTM with OP-LSTM, we see that the former yields the best performance in the 5-shot setting, whereas OP-LSTM outperforms LSTM in the 10-shot and 20-shot settings.

\subsection{Performance comparison on few-shot image classification} \label{sec:images}

\paragraph{Within-domain}

Next, we investigate the within-domain performance of OP-LSTM and LSTM on few-shot image classification problems, namely, Omniglot, miniImageNet, and CUB.
The results for the Omniglot dataset are displayed in Table~\ref{tab:omniglot}. 
Note that the LSTM has many more parameters than the other methods as it consists of multiple fully-connected layers with large hidden dimensions, which were found to give the best validation performance. 
As we can see, the plain LSTM (with batching) does not yield competitive performance compared with the other methods, in spite of the fact that it has many more parameters and, in theory, could learn any learning algorithm.
This shows that the LSTM is hard to optimize and struggles to find a good solution in more complex few-shot learning settings, i.e., image classification. 
OP-LSTM, on the other hand, which separates the learning procedure from the input representation, yields competitive performance compared with MAML and ProtoNet in both the 1-shot and 5-shot settings, whilst using fewer parameters than the plain LSTM.

\begin{table}[!htb]
    \centering
    \caption{The mean test accuracy (\%) on 5-way Omniglot classification across 3 different runs. The 95\% confidence intervals are displayed as $\pm x$, and calculated over all runs and meta-test tasks (600 per run). The plain LSTM is outperformed by MAML. All methods (except LSTM) used a fully-connected feed-forward classifier. The best performances are dislpayed in bold font.}
    \label{tab:omniglot}
    \begin{tabular}{lrrr}
    \toprule
        Technique & parameters &  1-shot & 5-shot \\
    \midrule
        MAML & $247\,621$ &  84.1 $\pm$ 0.90 & \textbf{93.5} $\pm$ 0.30  \\
        ProtoNet & $247\,621$ &  83.6 $\pm$ 0.88 & 93.4 $\pm$ 0.29 \\
        \midrule
        LSTM & $13\,530\,097$ & 72.6 $\pm$ 0.90 & 84.8 $\pm$ 0.50 \\
        OP-LSTM (ours) & $249\,167$  &\textbf{84.3} $\pm$ 0.90 & 91.8 $\pm$ 0.30 \\
        \bottomrule
    \end{tabular}
\end{table}

The results for miniImageNet and CUB are displayed in Table~\ref{tab:minandcub}.
Note that again, the LSTM uses more parameters than other methods as it consists of multiple large fully-connected layers which were found to yield the best validation performance. 
Nonetheless, it is applied on top of representations computed with the Conv-4 backbone, which is also used by all other methods. 
As we can see, the plain LSTM approach performs at chance level, again suggesting that the optimization problem of finding a good learning algorithm is too complex for this problem. 
The OP-LSTM, on the other hand, yields competitive or superior performance compared with all tested baselines on both miniImageNet and CUB, regardless of the number of shots, which shows the advantage of decoupling the input representation from the learning procedure.

\begin{table}[!htb]
\centering
\caption{Meta-test accuracy scores on 5-way miniImageNet and CUB classification over 3 runs. The 95\% confidence intervals are displayed as $\pm$ x, and calculated over all runs and meta-test tasks (600 per run). All methods used a Conv-4 backbone as a feature extractor. The ``-" indicates that the method did not finish within 2 days of running time. The best performances are dislpayed in bold font.}
\label{tab:minandcub}
\begin{tabular}{lrrrrr}
\toprule
& & \multicolumn{2}{c}{miniImageNet} & \multicolumn{2}{c}{CUB} \\
\cmidrule(lr){3-4} \cmidrule(lr){5-6}
Technique & parameters & 1-shot & 5-shot & 1-shot & 5-shot \\
 \midrule
MAML & $121\,093$ & 48.6 $\pm$ 1.04 & 63.0 $\pm$ 0.54 & 57.5 $\pm$ 1.04 & \textbf{74.8} $\pm$ 0.51 \\
Warp-MAML & $231\,877$ &  50.4 $\pm$ 1.04 & 65.6 $\pm$  0.53 & 59.6 $\pm$  1.00 & 74.2 $\pm$  0.51 \\
SAP & $412\,852$ & \textbf{53.0} $\pm$ 1.08 &  67.6 $\pm$  0.51 & \textbf{63.5} $\pm$ 1.00 & 73.9 $\pm$ 0.51 \\
ProtoNet & $121\,093$ & 50.1 $\pm$ 1.04 & 65.4 $\pm$ 0.53 & 50.9 $\pm$ 1.01 & 63.7 $\pm$ 0.55 \\
\midrule
LSTM & $55 \,879\,349$ &  20.2 $\pm$ 0.20 & 19.4 $\pm$ 0.20 & - & -\\
OP-LSTM (ours) & $141\,187$ & 51.9 $\pm$ 1.04 & \textbf{67.9} $\pm$ 0.50 & 60.2 $\pm$ 1.04 & 73.1 $\pm$ 0.52 \\
\bottomrule
\end{tabular}
\end{table}

\paragraph{Cross-domain}

Next, we investigate the cross-domain performance of the LSTM and OP-LSTM, where the test tasks come from a different a different dataset than the training tasks. 
We test this in two scenarios: train on miniImageNet and evaluate on CUB (MIN $\rightarrow$ CUB) and vice verse (CUB $\rightarrow$ MIN). 
The results of this experiment are displayed in Table~\ref{tab:crossdomain64channels}. 
Again, the plain LSTM does not outperform a random classifier, whilst the OP-LSTM yields superior performance in every tested scenario, showing its versatility in this challenging setting.  

\begin{table}[!htb]
\centering
\caption{Average cross-domain meta-test accuracy scores over 5 runs using a Conv-4 backbone. Techniques trained on tasks from one data set and were evaluated on tasks from another data set. The 95\% confidence intervals are displayed as $\pm$ x, and calculated over all runs and meta-test tasks. The ``-" indicates that the method did not finish within 2 days of running time. The best performances are dislpayed in bold font.}
\label{tab:crossdomain64channels}
\begin{tabular}{lcccccc}
\toprule
& \multicolumn{2}{c}{MIN $\rightarrow$ CUB} & \multicolumn{2}{c}{CUB $\rightarrow$ MIN} \\
\cmidrule(lr){2-3} \cmidrule(lr){4-5}
 & 1-shot & 5-shot  & 1-shot & 5-shot \\ 
\midrule
MAML &  37.9 $\pm$ 0.40 & 53.6 $\pm$ 0.40 & 31.1 $\pm$ 0.36 & 45.8 $\pm$ 0.39 \\
Warp-MAML & 42.0 $\pm$ 0.43 & 56.9 $\pm$ 0.42 & 31.1 $\pm$ 0.35 & 41.3 $\pm$ 0.36 \\
SAP  & 41.5 $\pm$ 0.44 & 58.0 $\pm$ 0.41 & 33.3 $\pm$ 0.39 & 47.1 $\pm$ 0.39 \\
ProtoNet & 39.7 $\pm$ 0.41 & 56.0 $\pm$ 0.41 & 31.7 $\pm$ 0.34 & 45.3 $\pm$ 0.38 \\
\midrule
LSTM & 20.1 $\pm$ 0.28 & 20.0 $\pm$ 0.25 & - & -\\
OP-LSTM (ours) & \textbf{42.3} $\pm$ 0.42 & \textbf{58.5} $\pm$ 0.41 & \textbf{35.8} $\pm$ 0.40 & \textbf{49.0} $\pm$ 0.40 \\
\bottomrule
\end{tabular}
\end{table}

\subsection{Analysis of the learned weight updates}\label{sec:howupdate}

Lastly, we investigate how OP-LSTM updates the weights of the base-learner network. 
More specifically, we measure the cosine similarity and Euclidean distance between the OP-LSTM updates and updates made by gradient descent or Prototypical network. 
Denoting the initial final classifier weight matrix as $\mathbf{H}^{(L)}_0$, the OP-LSTM update direction after $T$ updates is $\Delta_{\mathit{OP}} = \vec{\mathbf{H}}^{(L)}_T - \vec{\mathbf{H}}^{(L)}_0$, where $\vec{\mathbf{M}}$ means that we vectorize the matrix by flattening it.
Similarly, we can measure the update compared with the initial weight matrix and those obtained by employing nearest-prototype classification ($\mathbf{H}^{(L)}_{\mathit{Proto}}$) as done in Prototypical network or gradient descent  $\mathbf{H}^{(L)}_{\mathit{GD}}$, where the latter is obtained by performing $T$ gradient update steps (with a learning rate of 0.01). 
These updates are associated with the update direction vectors $\Delta_{\mathit{Proto}} = \vec{\mathbf{H}}^{(L)}_\mathit{Proto} - \vec{\mathbf{H}}^{(L)}_0$ and $\Delta_{\mathit{GD}} = \vec{\mathbf{H}}^{(L)}_{GD} - \vec{\mathbf{H}}^{(L)}_0$.
We can then measure the distance between the update direction $\Delta_{\mathit{OP}}$ of the OP-LSTM and $\Delta_{\mathit{Proto}}$ and $\Delta_{\mathit{GD}}$.
As a distance measure, we use the Euclidean distance.
In addition, we also measure the cosine similarity between the update directions as an inverse distance measure that is invariant to the scale and magnitudes of the vectors. 
After every $2\,500$ episodes, we measure these Euclidean distances and cosine similarity scores on the validation tasks, and average the results over 3 runs. 

\begin{figure}[!htb]
    \centering
    \begin{subfigure}{0.48\textwidth}
    \includegraphics[width=\textwidth]{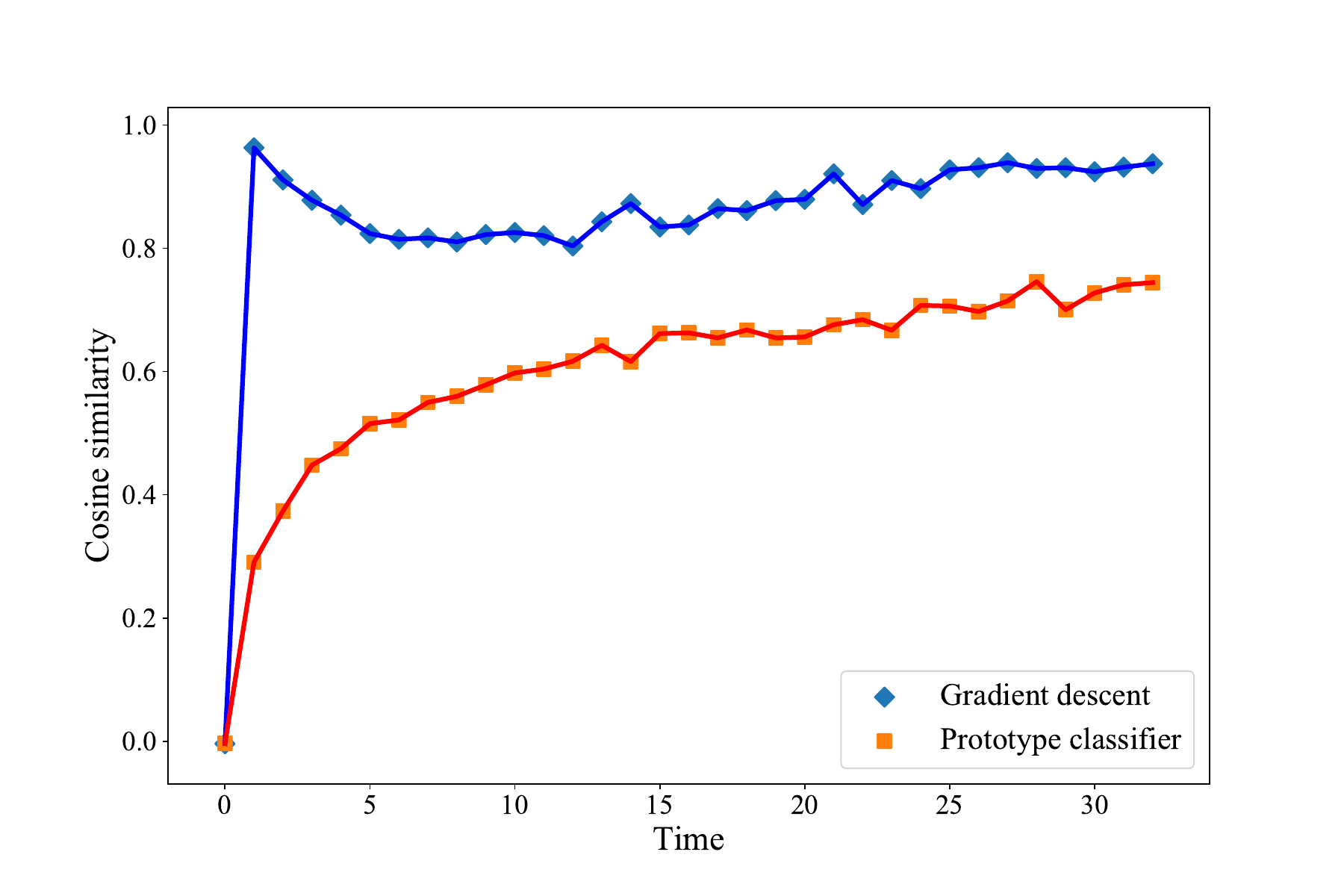}
    \caption{Cosine similarity}
    \end{subfigure}
    \begin{subfigure}{0.48\textwidth}
    \includegraphics[width=\textwidth]{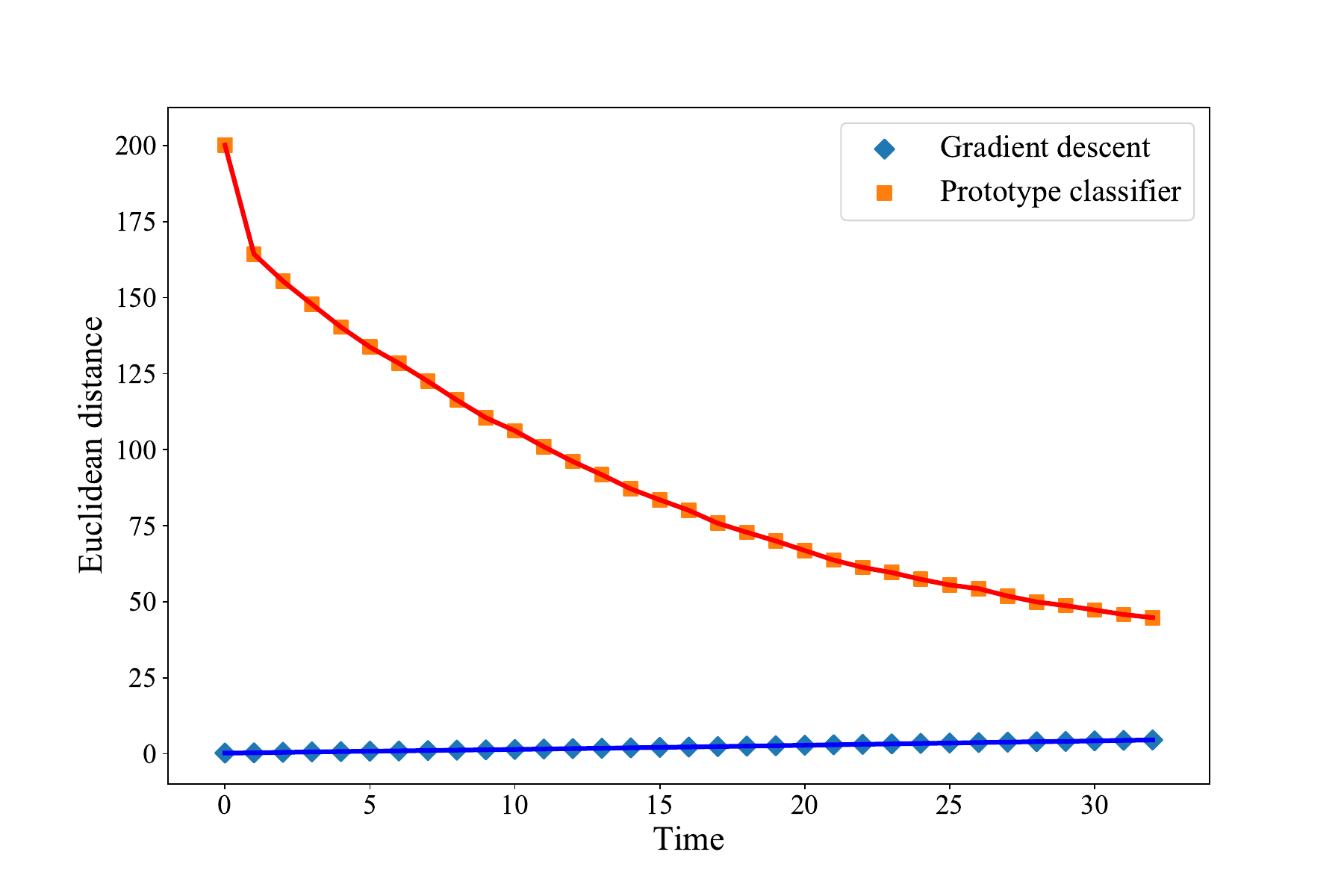}
    \caption{Euclidean distance}
    \end{subfigure}
    \caption{The average cosine similarity (left) and Euclidean distance (right) between the weight update directions of the OP-LSTM and a prototype-based and gradient-based classifier as a function of time on 5-way 1-shot miniImageNet classification. Each point on the x-axis indicates a validation step, which is performed after every $2\,500$ episodes. The results are averaged over 3 runs with different random seeds and the 95\% confidence intervals are shown as shaded regions. The confidnce intervals are within the size of the symbols and imperceptible. As time progresses, the updates performed by OP-LSTM become more similar to those of gradient descent and prototype-based classifiers (increasing cosine similarity).}
    \label{fig:cosine-euclid}
\end{figure}

The results of this experiment are displayed in Figure~\ref{fig:cosine-euclid}. 
As we can see, the cosine similarity between the weight update directions of OP-LSTM and gradient descent and prototype-based classifiers increases with training time. 
OP-LSTM very quickly learns to update the weights in a similar direction as gradient descent, followed by a gradual decline in similarity, which is later followed by a gradual increase. 
This gradual decline may be to incorporate more prototype-based updates. 
Looking at the Euclidean distance, we observe the same pattern for the similarity compared with the prototype-based classifier, as the distance between the updates decreases (indicating a higher similarity). 
The Euclidean distance between OP-LSTM updates and gradient updates slightly increase over time, which may be a side effect of the sensitivity to scale and magnitude of this distance measure. 
Thus, even if both would perform gradient descent, but with different learning rates, 
The cosine similarity gives a better idea of directional similarity as it abstracts away from the magnitude of the vectors.

\section{Relation to other methods}\label{sec:relation}

Here, we study the relationship of OP-LSTM to other existing meta-learning methods. 
More specifically, we aim to show that OP-LSTM is a general meta-learning approach, which can {\it approximate} the behaviour of different classes of meta-learning, such as optimization-based meta-learners (e.g., MAML, \citet{finn2017model}) and metric-based methods (e.g., Prototypical network,  \citet{snell2017prototypical}).

\paragraph{Model-agnostic meta-learning (MAML)} 
MAML \citep{finn2017model} aims to learn good initialization parameters for a base-learner network $\mathbf{\theta} = \{ \mathbf{W}^{(1)}_0, \mathbf{W}^{(2)}_0, \ldots, \mathbf{W}^{(L)}_0, \bias^{(1)}_0, \bias^{(2)}_0, \ldots, \bias^{(L)}_0  \}$ such that new tasks can be learned efficiently using a few gradient update steps. 
Here, $\mathbf{W}^{(\ell)}_0$ is the initial weight matrix of layer $\ell$ and $\bias^{(\ell)}_0$ the initial bias vector of layer $\ell$ when presented with a new task.
 
The initial 2D hidden states $\mathbf{H}^{(\ell)}_0$ in OP-LSTM can be viewed as the initial weights $\mathbf{W}^{(\ell)}_0$ of the neural network in MAML. 
In MAML, the weights in layer $\ell$ for a given input are updated as $\weight^{(\ell)}_{t+1} = \weight^{(\ell)}_t - \eta \delta^{(\ell)} (\mathbf{p}^{(\ell - 1)}(\inp))^T$, where $\mathbf{a}^{(i)}(\inp)$ (with $1 \le i \le L$) is the vector of post-activation values in layer $i$ as a result of the input $\inp$, and $\delta^{(i)} = \nabla_{\mathbf{a}^{(i)}} \mathcal{L} (\inp)$, where $\mathcal{L} (\inp, \out)$ is the loss on input $\inp$ given the ground-truth target $\out$, and $\eta$ is the learning rate.

Instead of using this hand-crafted weight update rule, OP-LSTM learns the update rule using the outer product of LSTM hidden states and the input activation.
From Eq.~\ref{eq:updateH} it follows that OP-LSTM is capable of updating the weights $\mathbf{H}^{(\ell)}_t$ with gradient descent by setting $\hid^{(\ell)}_{t+1}(\inp) = -\eta \delta^{(\ell)} = -\eta \nabla_{\mathbf{a}^{(\ell)}} \mathcal{L} (\inp, \out)$. 
(Note that in Eq.~\ref{eq:updateH} the gradient is also normalized by the Frobenius norm, which is formally not part of MAML.)
We note that the inputs to the coordinate-wise LSTM contain the necessary information to compute the errors $\delta^{(\ell)}$ in every layer.
That is, for the output layer, the LSTM receives the ground-truth output and prediction in the output layer.
For earlier layers, the LSTM receives the backpropagated messages (the errors), as well as the activations. 
Consequently, OP-LSTM can update the 2D hidden states $\mathbf{H}^{(\ell)}$ with gradient descent, as MAML. 
OP-LSTM is thus an approximate generalization of MAML as it could learn to perform the same weight matrix updates, although OP-LSTM does not update the bias vectors given a task.

\paragraph{Prototypical network}
Prototypical network \citep{snell2017prototypical} aims to learn good initial weights $\mathbf{\theta} = \{ \mathbf{W}^{(1)}_0, \mathbf{W}^{(2)}_0, \ldots, \mathbf{W}^{(L-1)}_0, \bias^{(1)}_0, \bias^{(2)}_0, \ldots, \bias^{(L-1)}_0  \}$ for all parameters except for the final layer, such that a nearest-prototype classifier yields good performance. 
Let $f_{\theta}(\inp_i)$ be the embeddings produced by this $(L-1)$-layered network for a given observation $\inp_i$ (from the support set).
Note that the network has $(L-1)$ layers as this is the feature embedding module without the output layer. 
Prototypical network computes centroids $\mathbf{c}_n = \frac{1}{|X_n|} \sum_{\inp_i \in X_n} f_{\theta}(\inp_i)$ for every class $n$, where $X_n$ is the set of all support inputs with ground-truth class $n$, and $f_{\theta}(\inp)$ is the embedding of input $\inp$. 
Then, the predicted score of a new input $\hat{\inp}$ for class $n$ is then given by $\hat{y}_n(\hat{\inp}) = \frac{\exp (-d(f_{\theta}(\hat{\inp}), \mathbf{c}_n))}{\sum_{n'} \exp (-d(f_{\theta}(\hat{\inp}), \mathbf{c}_{n'}))}$, where $d(\inp_i, \inp_j) = || \inp_i - \inp_j ||_2^2$ is the squared Euclidean distance, and $n'$ is a variable iterating over all classes.

This nearest-prototype classifier can be seen as a regular linear output layer \citep{triantafillou2020meta}. 
To see this, note that the prediction score for class $j$ is given by
\begin{align}
    \hat{y}_j(\hat{\inp}) = || f_{\theta} (\hat{\inp}) - \mathbf{c}_n  ||^2_2 &= (f_{\theta} (\hat{\inp}) - \mathbf{c}_n )^T(f_{\theta} (\hat{\inp}) - \mathbf{c}_n ) \\
    &= f_{\theta} (\hat{\inp})^T f_{\theta} (\hat{\inp}) - 2 f_{\theta} (\hat{\inp})^T   \mathbf{c}_n +  \mathbf{c}_n^T  \mathbf{c}_n \\
    &\propto - 2 f_{\theta} (\hat{\inp})^T   \mathbf{c}_n +  \mathbf{c}_n^T  \mathbf{c}_n,
\end{align}
where we ignored the first term ($f_{\theta} (\hat{\inp})^T f_{\theta} (\hat{\inp})$) as it is constant across all classes $n$. 
The prediction score for class $j$ is thus obtained by taking the dot product between the input embedding $f_{\theta}(\hat{\inp})$ and $-2\mathbf{c}_n$ and by adding a bias term $b_n = \mathbf{c}^T_n \mathbf{c}_n$.
Thus, the prototype-based classifier is equivalent to a linear output layer, i.e., $\hat{\inp} = \weight^{(L)} f_{\theta}(\hat{\inp}) + \bias^{(L)}$ where the $n$-th row of $\weight^{(L)}$ corresponds to $-2\mathbf{c}_n$ and the $n$-th element of $\bias^{(L)}$ is equal to $\mathbf{c}^T_n \mathbf{c}_n$.
OP-LSTM can approximate the behavior of Prototypical network with $T=1$ steps per task as follows.
First, assume that the underlying base-learner network is the same for Prototypical network and OP-LSTM, i.e., the initialization of the hidden state is equivalent to the initial weights of the base-learner used by Prototypical network $\mathbf{H}_0^{(\ell)} = \weight^{(\ell)}_0$ for $\ell \in \{1,2,\ldots,L-1\}$, and that the hidden state of the output layer in OP-LSTM is a matrix of zeros, i.e., $\mathbf{H}^{(L)}_0 = \mathbf{0}$.  
Second, let the hidden states of the LSTM in OP-LSTM be be a vector of zeros $\hid^{(\ell)} = \mathbf{0}$ for every layer $\ell < L$, and let the hidden state of the output layer given the example $\mathbf{x}'_i = (\mathbf{x}_i, \mathbf{y}_i)$ be the label identity function $\hid^{(L)}(\mathbf{x}'_i) = \out_i$ (which can be learned by an LSTM).
Then, OP-LSTM will update the hidden states as follows using Eq.~\ref{eq:updateH}. 
The $n$-th row of $\mathbf{H}^{(L)}$ will equal $\gamma \frac{1}{M}\sum_{\inp_i \in X_n} \frac{\act^{(L-1)}(\inp_i)}{|| \act^{(L-1)}(\inp_i) ||_F}$, where $X_n = \{ \inp_i \in \Dtr | \out_i=\mathbf{e}_n \}$ is the set of training inputs with class $n$, and $\gamma$ and $M$ are the learning rate of OP-LSTM and number of examples respectively. 
Note that this expression corresponds to the scaled prototype (mean of the embeddings) of class $n$, that is, $\gamma \bar{\mathbf{c}}_n$, where $\bar{\mathbf{c}}_n = \frac{1}{M} \sum_{\inp_i \in X_n} \frac{\act^{(L-1)}(\inp_i)}{|| \act^{(L-1)}(\inp_i) ||_F}$.
The prediction for the $n$-th class for a given input $\hat{\mathbf{x}}$ is thus given by $\gamma \bar{\mathbf{c}}_n^T \act^{(L-1)}(\inp) + b_n^{(L)}$, where we omitted the time step for $\act^{(L-1)}$ and $b_n$ is a fixed bias in the output layer.  
Note that for $\gamma=-2$, the first term ($-2\bar{\mathbf{c}}_n^T\act^{(L-1)}(\inp)$) resembles the first term in the prediction made by Prototypical network for class $n$, which is given by $-2 \mathbf{c}_n^T \act^{(L-1)}(\inp)$, where $\act^{(L-1)}(\inp) = f_{\theta}(\inp)$.
Hence, OP-LSTM can learn to approximate (up to the bias term) a normalized Prototypical network classifier.

We have thus shown that OP-LSTM can learn to implement a parametric learning algorithm (gradient descent) as well as a non-parametric learning algorithm (prototype-based classifier), demonstrating the flexibility of the approach.

\section{Conclusions}

Meta-learning is a strategy to enable deep neural networks to learn from small amounts of data. 
The field has witnessed an increase in popularity in recent years, and many new techniques are being developed. 
However, the potential of some of the earlier techniques have not been studied thoroughly, despite promising initial results.
In our work, we revisited the plain LSTM approach proposed by \citet{hochreiter2001learning} and \citet{younger2001meta}. This approach simply ingests the training data for a given task, and conditions the predictions of new query inputs on the resulting hidden state.

We analysed this approach from a few-shot learning perspective and uncovered two potential issues for embedding a learning algorithm into the weights of the LSTM: 1)~the hidden embeddings of the support set are not permutation invariant, and 2)~the learning algorithm and the input embedding mechanism are intertwined, which leads to a challenging optimization problem and an increased risk of overfitting. 
In our work, we proposed to overcome issue 1) by mean pooling the embeddings of individual training examples, rendering the obtained embedding permutation invariant. 
We found that this method is highly effective and increased the performance of the plain LSTM on both few-shot sine wave regression and image classification. 
Moreover, with this first solution, the plain LSTM approach already outperformed the popular meta-learning method MAML \citep{finn2017model} on the former problem.
It struggled, however, to yield good performance on few-shot image classification problems, highlighting the difficulty of optimizing this approach.   

In order to resolve this difficulty, we proposed a new technique, Outer Product LSTM (OP-LSTM), that uses an LSTM to update the weights of a base-learner network.
By doing this, we effectively decouple the learning algorithm (the weight updates) from the input representation mechanism (the base-learner), solving issue 2), as done in previous works \citep{ravi2017optimization, andrychowicz2016learning}.  
Compared with previous works, OP-LSTM does not receive gradients as inputs.
Our theoretical analysis shows that OP-LSTM is capable of performing an approximate form of gradient descent (as done in MAML \citep{finn2017model}) as well as a nearest prototype based approach (as done in Prototypical network \citep{snell2017prototypical}), showing the flexibility and expressiveness of the method. 
Empirically, we found that OP-LSTM overcomes the optimization issues associated with the plain LSTM approach on few-shot image classification benchmarks, whilst using fewer parameters. 
It yields competitive or superior performance compared with MAML \citep{finn2017model} and Prototypical network \citep{snell2017prototypical}, both of which it can approximate.

\paragraph{Future work}
When the base-learner is a convolutional neural network, we applied OP-LSTM on top of the convolutional feature embeddings. 
A fruitful direction for future research would be to propose a more general form of OP-LSTM that can update also the convolutional layers. 
This would require new backward message passing protocols to go through pooling layers often encountered in convolutional neural networks. 

Moreover, we note that OP-LSTM is one way to overcome the two issues associated with the plain LSTM approach, but other approaches could also be investigated. 
For example, one could try to implement a convLSTM \citep{shi2015convolutional} such that the LSTM can be applied directly to raw inputs, instead of only after the convolutional backbone in case of image classification problems.

Another fruitful direction for future work would be to investigate different recurrent neural architectures and their ability to perform meta-learning. 
In the pioneering work of \citet{hochreiter2001learning} and \citet{younger2001meta}, it was shown that only LSTMs were successful whereas vanilla recurrent neural networks and Elman networks failed to meta-learn simple functions. 
It would be interesting to explore how architectural design choices influence the ability of recurrent networks to perform meta-learning. 

Lastly, OP-LSTM is a method to learn the weight update rule for a base-learner network, and is thus orthogonal to many advances and new methods in the field of meta-learning, such as Warp-MAML \citep{Flennerhag2020Meta-Learning} and SAP \citep{huisman2023}.
Since this is a nontrivial extension of these methods, we leave this for future work. 
We think that combining these methods could yield new state-of-the-art performance.

{\bf Acknowledgments}

This work was performed using the compute resources from the Academic Leiden Interdisciplinary Cluster Environment (ALICE) provided by Leiden University.

\section*{Declarations}

\subsection*{Funding}
Not applicable: no funding was received for this work.

\subsection*{Conflict of interest} All authors certify that they have no affiliations with or involvement in any organization or entity with any financial interest or non-financial interest in the subject matter or materials discussed in this manuscript.

\subsection*{Ethics approval}
Not applicable. 

\subsection*{Consent to participate}
Not applicable.

\subsection*{Consent for publication}

Not applicable: this research does not involve personal data, and publishing of this manuscript will not result in the disruption of any individual's privacy.

\subsection*{Availability of data and material}

All data that was used in this research have been published as benchmarks by \citet{deng2009imagenet,vinyals2016matching} (miniImageNet) and \citet{wah2011caltech} (CUB), and is publicly available. The data generator for sine wave regression experiments can be found in the provided code (see below).

\subsection*{Code availability}
All code that was used for this research is made publicly available at \url{https://github.com/mikehuisman/lstm-fewshotlearning-oplstm}.

\subsection*{Authors' contributions}

MH has conducted the research presented in this manuscript.
TM, AP, and JvR have regularly provided feedback on the work, contributed towards the interpretation of results, and have critically revised the whole.    

All authors approve the current version to be published and agree to be accountable for all aspects of the work in ensuring that questions related to the accuracy or integrity of any part of the work are appropriately investigated and resolved.

\subsection*{Employment} All authors declare that there is no recent, present, or anticipated employment by any organization that may gain or lose financially through the publication of this manuscript.

{
\bibliographystyle{abbrvnat}
\bibliography{arxiv}
}

\appendix

\section{Sine wave regression: additional results} \label{sec:sinewave-appendix}

We also performed an experiment to investigate the effect of the input representation on the performance of the plain LSTM approach (proposed by \citet{younger2001meta, hochreiter2001learning}) on the 5-shot sine wave regression performance.
The experimental setting follows the setup described in Section~\ref{sec:perminvariance}.
For every input format, we performed hyperparameter tuning with the same randomly sampled hyperparameter configurations using Table~\ref{tab:randomsearch}. 
The performances of the best validated models per input format are displayed in Table~\ref{tab:lstminputs}.
The best performance is obtained by feeding the current input, previous target, and the previous prediction into the LSTM, although the differences with other inputs are small. 

\begin{table}[!htb]
    \centering
    \caption{The influence of different input information on the performance of the LSTM on 5-shot sine wave regression. 95\% confidnce intervals are displayed as $\pm x$. The best performances are dislpayed in bold font.}
    \label{tab:lstminputs}
    \begin{tabular}{ccccc}
    \toprule
         Input $\mathbf{x}_t$ & Prev target $y_{t-1}$ & Prev pred $\hat{y}_{t-1}$ & Prev error $e_{t-1}$ & 5-shot MSE  \\
         \midrule
         \checkmark & \checkmark & & & 0.04 $\pm$ 0.002 \\
         \checkmark & \checkmark & \checkmark & & \textbf{0.03} $\pm$ 0.002 \\
         \checkmark & \checkmark & & \checkmark &  0.05 $\pm$ 0.004 \\
         \checkmark & \checkmark & \checkmark & \checkmark & 0.06 $\pm$ 0.011 \\
         \bottomrule
    \end{tabular}
\end{table}

\section{Hyperparameter tuning} \label{sec:hypers}

\subsection{Permutation invariance experiments}

For the permutation invariance experiments on few-shot sine wave regression, we sampled 20 random configurations for the plain LSTM from the distributions displayed in Table~\ref{tab:randomsearch} and validated their performance on 5-shot ($k=5$) sine-wave regression.
We selected the best configuration and evaluated it on the meta-test tasks,

\begin{table}[!htb]
    \centering
    \caption{The used ranges and distributions for tuning the hyperparameters with random search for sine wave regression. }
    \label{tab:randomsearch}
    \begin{tabular}{ll}
    \toprule
        Hyperparameter & Range \\
        \midrule
        Number of layers & Uniform(\{1,2,3,4\})  \\
        Hidden dimensions & Uniform(\{1,3,8,20,40\}) \\
        Meta-batch size & Uniform(\{1,2,3,4\}) \\
        Learning rate & LogUniform(1e-5, 4e-2) \\
        Unroll steps &  Uniform(\{1,2,\ldots,14\}) \\
        \bottomrule
    \end{tabular}
\end{table}

For Omniglot, we performed random search with a function evaluation budget of 100, with a fixed learning rate of 0.001. 
The architecture of the plain LSTM with sequential data processing was sampled uniformly at random from \{1024-512-256-128-64,2048-1024-512-128-64,2048-1024-512-256-128,1024-600-400-200-92,1024-512-512-256-128-64,1024-512-512-256-256-128-64,612-400-256-128-64,1024-1024-1024-512-256-128-64,2048-1024-512-180-100,1024-580-280-160-80,256-128-64,512-256-128-64,128-64-64-64,256-128-64,512-256-64,256-128-100,128-64-64-64-64,64-64-64-64,50-50\}, the number of passes over the support data $T$ was sampled uniformly at random from $\{1,2,\ldots,10\}$, and the meta-batch size from $\{1,2,\ldots,32\}$. 
We used the best hyperparameter configuration of the sequential plain LSTM for the plain LSTM with batching to compare the differences in performance. 

\subsection{Omniglot}\label{sec:omniglot}

For the \textbf{plain LSTM} approach, we used the best hyperparameter configuration found for the permutation invariance experiments. 

For \textbf{OP-LSTM}, we performed a grid search, varying the meta-batch size within \{1,4,8,16,32\}, the architecture of the coordinate-wise LSTM within \{20-1, 10-10-1, 40-5, 40-20-1, 20-20-20-5\} (note that the last element is always 1 because it operates per coordinate), and the number of passes over the support set within \{1,3,5,10\}.

\begin{figure}
    \centering
    \includegraphics[width=\textwidth]{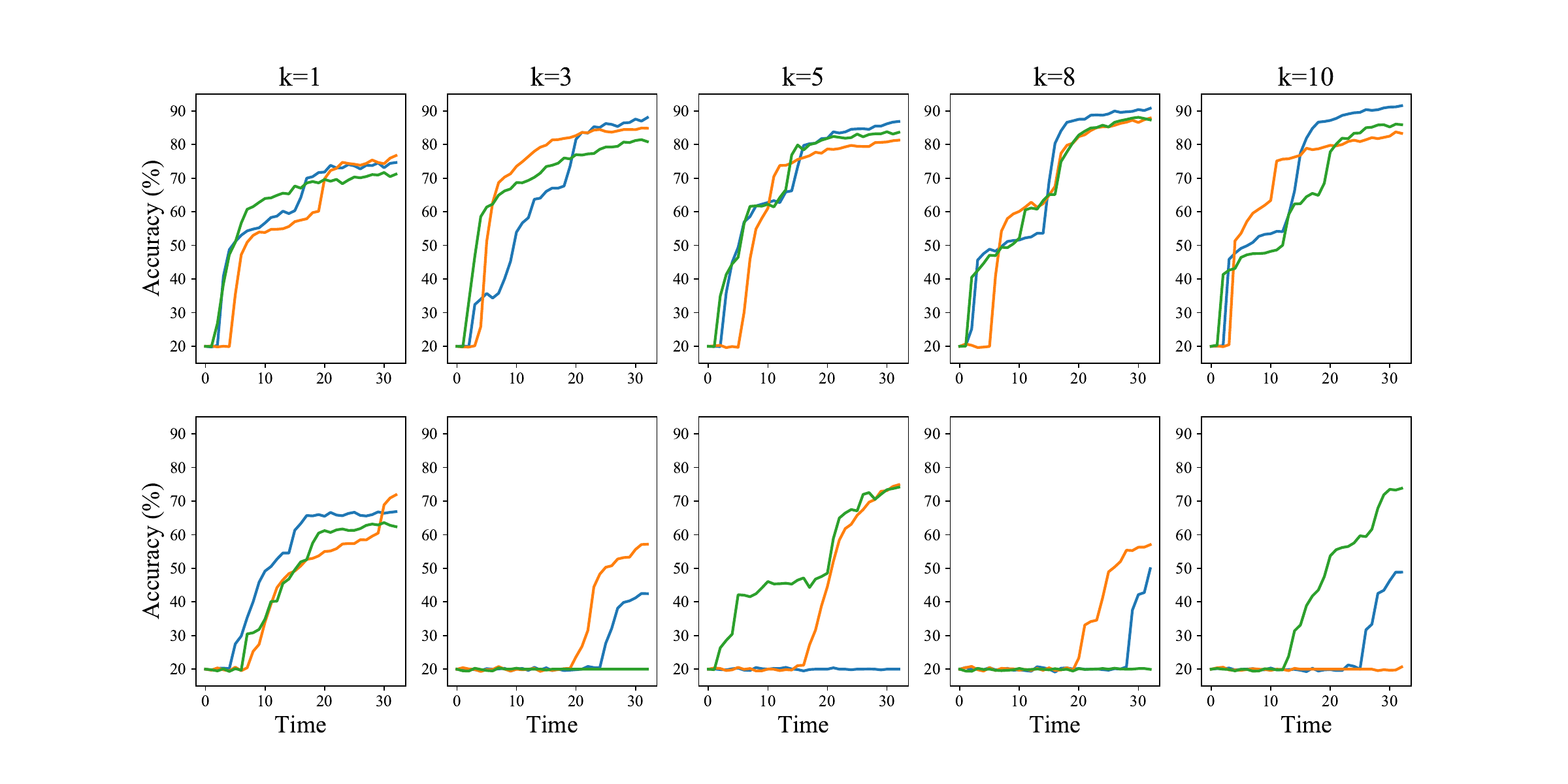}
    \caption{The mean validation accuracy of the LSTM over time on Omniglot for every of the three different runs, for different numbers of examples per class $k$. \textbf{Top row:} LSTM with batching (mean-pooling). \textbf{Bottom row:} LSTM with sequential data ingestion. As we can see, batching improves the stability of the training process.}
    \label{fig:omnidetailed}
\end{figure}

\paragraph{Detailed learning curves for the plain LSTM on Omniglot} Here, we show the validation learning curves of the sequential LSTM and the LSTM which uses batching to complement the results displayed in Section~\ref{sec:perminvariance}. 
Figure~\ref{fig:omnidetailed} displays the validation learning curves of the LSTM with batch data ingestion (top row) and the LSTM with sequential data processing (bottom row). 
As we can see, batching increases the stability of the training process and makes the LSTM less sensitive to the random initialization, as every run succeeds to reach convergence in contrast to the sequential LSTM. 

\subsection{miniImageNet and CUB}\label{sec:runningtimes}

For \textbf{plain LSTM}, we used random search with a budget of 130 function evaluations, the meta-batch size was sampled uniformly between 1 and 48, the number of layers between 1 and 4, the hidden size log-uniformly between 32 and 3200, and the number of passes T over the support dataset uniformly between 2 and 9.

For \textbf{OP-LSTM}, we performed the same grid search as for Omniglot. 
We use the best found hyperparameters for both methods on miniImageNet also on CUB.

We also measured the running times of the techniques on miniImageNet and CUB, as shown in Table~\ref{tab:runningtimes}. 
We note that the running times may be affected by the server's load and thus can only give a rough estimation of the required amount of compute time. 
As we can see, the plain LSTM is the slowest method, despite achieving random performance on miniImageNet. 
OP-LSTM, in contrast, is more efficient. 

\begin{table}[!htb]
\centering
\caption{Mean running times on 5-way miniImageNet and CUB classification over 3 runs. All methods used a Conv-4 backbone as a feature extractor. ``$x$h$y$min'' means $x$ hours and $y$ minutes. The ``-" indicates that the method did not finish within 2 days of running time.}
\label{tab:runningtimes}
\begin{tabular}{llllll}
\toprule
& &\multicolumn{2}{c}{miniImageNet} & \multicolumn{2}{c}{CUB} \\
\cmidrule(lr){3-4} \cmidrule(lr){5-6} 
Technique & params & 1-shot & 5-shot& 1-shot & 5-shot\\
 \midrule
MAML & $121\,093$  & 13h9min & 12h1min & 26h57min & 17h39min     \\
Warp-MAML & $231\,877$ &  12h25min & 12h30min & 13h6min & 12h48min          \\
SAP & $412\, 852$ &  5h40min & 11h14min & 7h11min &  11h17min                \\
ProtoNet & $121\, 093$ & 4h14min & 5h6min & 31h18min & 38h46min         \\
\midrule
LSTM & $55\, 879\, 349 $ & 40h14min & 46h47min &  - & -            \\
OP-LSTM (ours) & $141\, 187$ & 4h50min & 5h31min & 31h58min & 40h8min    \\
\bottomrule
\end{tabular}
\end{table}

\subsection{Robustness to random seeds}

Here, we investigate the robustness of the investigated methods to the random seed for the few-shot image classification experiments performed in Section~\ref{sec:images}.
We perform th
Instead of computing the confidence intervals over the performances of all test tasks for all seeds, we now compute the confidence interval over the mean test performance per run. 
As we perform three runs per method, we compute the confidence intervals over three observations per method. 
Note that the mean performance does not change as taking the mean of the three means will be equivalent (as the means are based on an equal number of task performances). 

\subsubsection{Within-domain}
Here, we present additional results for the conducted within-domain image classification experiments.

\paragraph{Omniglot}
The mean test performance and confidence intervals over the random seeds for Omniglot image classification are shown in Table~\ref{tab:omniglot2}. 
As we can see, the confidence intervals are higher than in previous experiments because the intervals are computed over 3 observations instead of 1800 individual test task performances (600 per run). 
As we can see, the LSTM is unstable, supporting the hypothesis that the optimization problem is difficult. 
OP-LSTM, on the other hand, is less sensitive to the chosen random seed and has a stability that is comparable to that of MAML. 

\begin{table}[!htb]
    \centering
    \caption{The mean test accuracy (\%) on 5-way Omniglot classification across 3 different runs. The 95\% confidence intervals, computed over the mean performances of the 3 different random seeds, are displayed as $\pm x$. The plain LSTM is outperformed by MAML. All methods (except LSTM) used a fully-connected feed-forward classifier. The best performances are dislpayed in bold font.}
    \label{tab:omniglot2}
    \begin{tabular}{lccc}
    \toprule
        Technique & parameters &  1-shot & 5-shot \\
    \midrule
        MAML & $247\,621$ &  84.1 $\pm$ 3.10 & \textbf{93.5} $\pm$ 0.70  \\
        ProtoNet & $247\,621$ &  83.6 $\pm$ 0.52 & 93.4 $\pm$ 1.48 \\
        \midrule
        LSTM & $13\,530\,097$ & 72.6 $\pm$ 3.87 & 84.8 $\pm$ 6.12 \\
        OP-LSTM (ours) & $249\,167$  &\textbf{84.3} $\pm$ 3.18 & 91.8 $\pm$ 0.70 \\
        \bottomrule
    \end{tabular}
\end{table}

\paragraph{MiniImageNet and CUB}

The mean test performance and confidence intervals over the random seeds for miniImageNet and CUB image classification are shown in Table~\ref{tab:minandcub2}. 
In contrast to what we observed on Omniglot, the LSTM is now more stable.
This is caused by the fact that it consistently fails to learn a learning algorithm that performs better than random guessing, and thus performs stably at chance level. 

\begin{table}[!htb]
\centering
\caption{Meta-test accuracy scores on 5-way miniImageNet and CUB classification over 3 runs. The 95\% confidence intervals, computed over the mean performances of the 3 different random seeds, are displayed as $\pm$ x. All methods used a Conv-4 backbone as a feature extractor. The ``-" indicates that the method did not finish within 2 days of running time. The best performances are dislpayed in bold font.}
\label{tab:minandcub2}
\begin{tabular}{lccccc}
\toprule
& & \multicolumn{2}{c}{miniImageNet} & \multicolumn{2}{c}{CUB} \\
\cmidrule(lr){3-4} \cmidrule(lr){5-6}
Technique & params & 1-shot & 5-shot & 1-shot & 5-shot \\
 \midrule
MAML & $121\,093$ & 48.6 $\pm$ 4.00 & 63.0 $\pm$ 0.33 & 57.5 $\pm$ 0.83 & \textbf{74.8} $\pm$ 2.10  \\
Warp-MAML & $231\,877$ &  50.4 $\pm$ 2.58 & 65.6 $\pm$ 0.98 & 59.6 $\pm$ 2.15 & 74.2 $\pm$ 2.51  \\
SAP & $412\,852$ & \textbf{53.0} $\pm$ 3.71 &  67.6 $\pm$ 0.47 & \textbf{63.5} $\pm$ 6.24 & 73.9 $\pm$ 1.57 \\
ProtoNet & $121\,093$ & 50.1 $\pm$ 4.06 & 65.4 $\pm$ 2.84 & 50.9 $\pm$ 2.35 & 63.7 $\pm$ 0.47 \\
\midrule
LSTM & $55 \,879\,349$ &  20.2 $\pm$ 0.60 & 19.4 $\pm$ 0.47 & - & -\\
OP-LSTM (ours) & $141\,187$ & 51.9 $\pm$ 2.52 & \textbf{67.9} $\pm$ 2.40 & 60.2 $\pm$ 1.58 & 73.1 $\pm$ 1.57 \\
\bottomrule
\end{tabular}
\end{table}

\begin{table}[!htb]
\centering
\caption{Average cross-domain meta-test accuracy scores over 5 runs using a Conv-4 backbone. Techniques trained on tasks from one data set and were evaluated on tasks from another data set. The 95\% confidence intervals, computed over the mean performances of the 3 different random seeds, are displayed as $\pm$ x. The ``-" indicates that the method did not finish within 2 days of running time. The best performances are dislpayed in bold font.}
\label{tab:crossdomain64channels2}
\begin{tabular}{lcccccc}
\toprule
& \multicolumn{2}{c}{MIN $\rightarrow$ CUB} & \multicolumn{2}{c}{CUB $\rightarrow$ MIN} \\
\cmidrule(lr){2-3} \cmidrule(lr){4-5}
 & 1-shot & 5-shot  & 1-shot & 5-shot \\ 
\midrule
MAML &  37.9 $\pm$ 2.22 & 53.6 $\pm$ 0.67  & 31.1 $\pm$ 1.19 & 45.8 $\pm$ 2.06 \\
Warp-MAML & 42.0 $\pm$ 0.85 & 56.9 $\pm$ 4.16 & 31.1 $\pm$ 1.59 & 41.3 $\pm$ 1.37 \\
SAP  & 41.5 $\pm$ 3.72 & 58.0 $\pm$ 1.79 & 33.3 $\pm$ 2.33 & 47.1 $\pm$ 1.28 \\
ProtoNet & 39.7 $\pm$ 4.11 & 56.0 $\pm$ 4.89 & 31.7 $\pm$ 0.20 & 45.3 $\pm$ 1.84 \\
\midrule
LSTM & 20.1 $\pm$ 0.77 & 20.0 $\pm$ 0.40 & - & -\\
OP-LSTM (ours) & \textbf{42.3} $\pm$ 1.90 & \textbf{58.5} $\pm$ 1.49 & \textbf{35.8} $\pm$ 2.98 & \textbf{49.0} $\pm$ 0.80 \\
\bottomrule
\end{tabular}
\end{table}

\subsubsection{Cross-domain}

Lastly, we compute the confidence intervals in cross-domain settings and display the results in Table~\ref{tab:crossdomain64channels2}.
Again, the LSTM is a stable random guesser.
The other algorithms are less stable, but do yield a better performance. 
We cannot observe a general pattern of stability in the sense that one algorithm is consistently more stable than others.

\end{document}